\documentclass[lettersize,journal]{IEEEtran}
\usepackage{amsmath,amsfonts}
\usepackage{algorithmic}
\usepackage{algorithm}
\usepackage{array}
\usepackage[caption=false,font=normalsize,labelfont=sf,textfont=sf]{subfig}
\usepackage{textcomp}
\usepackage{stfloats}
\usepackage{url}
\usepackage{verbatim}
\usepackage{graphicx}
\usepackage{cite}
\usepackage{pifont}
\usepackage{adjustbox}
\usepackage{multirow}
\usepackage{booktabs}
\usepackage{bbm, dsfont}
\usepackage{amssymb}
\usepackage{xcolor}
\usepackage{colortbl}
\usepackage{bbding}

\begin{document}


\title{RAW-Adapter: Adapting Pre-trained Visual Model to Camera RAW Images and A Benchmark}


\author{Ziteng Cui, Jianfei Yang, Tatsuya Harada
\thanks{Ziteng Cui is with the RCAST, The University of Tokyo, Japan.}
\thanks{Jianfei Yang is with the MARS Lab at Nanyang Technological University, Singapore}
\thanks{Tatsuya Harada is with the RCAST, The University of Tokyo and RIKEN AIP, Japan.}
\thanks{\Envelope \ Corresponding Author: Tatsuya Harada (harada@mi.t.u-tokyo.ac.jp)}}


\markboth{}%
{Shell \MakeLowercase{\textit{et al.}}: A Sample Article Using IEEEtran.cls for IEEE Journals}

\IEEEpubid{}

\maketitle

\begin{abstract}

In the computer vision community, the preference for pre-training visual models has largely shifted toward sRGB images due to their ease of acquisition and compact storage. However, camera RAW images preserve abundant physical details across diverse real-world scenarios. Despite this, most existing visual perception methods that utilize RAW data directly integrate image signal processing (ISP) stages with subsequent network modules, often overlooking potential synergies at the model level.
Building on recent advances in adapter-based methodologies in both NLP and computer vision, we propose RAW-Adapter, a novel framework that incorporates learnable ISP modules as input-level adapters to adjust RAW inputs. At the same time, it employs model-level adapters to seamlessly bridge ISP processing with high-level downstream architectures. Moreover, RAW-Adapter serves as a general framework applicable to various computer vision frameworks.

Furthermore, we introduce \textbf{RAW-Bench}, which incorporates 17 types of RAW-based common corruptions, including lightness degradations, weather effects, blurriness, camera imaging degradations, and variations in camera color response. Using this benchmark, we systematically compare the performance of RAW-Adapter with state-of-the-art (SOTA) ISP methods and other RAW-based high-level vision algorithms. Additionally, we propose a RAW-based data augmentation strategy to further enhance RAW-Adapter’s performance and improve its out-of-domain (OOD) generalization ability. Extensive experiments substantiate the effectiveness and efficiency of RAW-Adapter, highlighting its robust performance across diverse scenarios.

\end{abstract}


\section{Introduction}
\label{sec1:intro}


\begin{figure}
    \centering
    \includegraphics[width=1.0\linewidth]{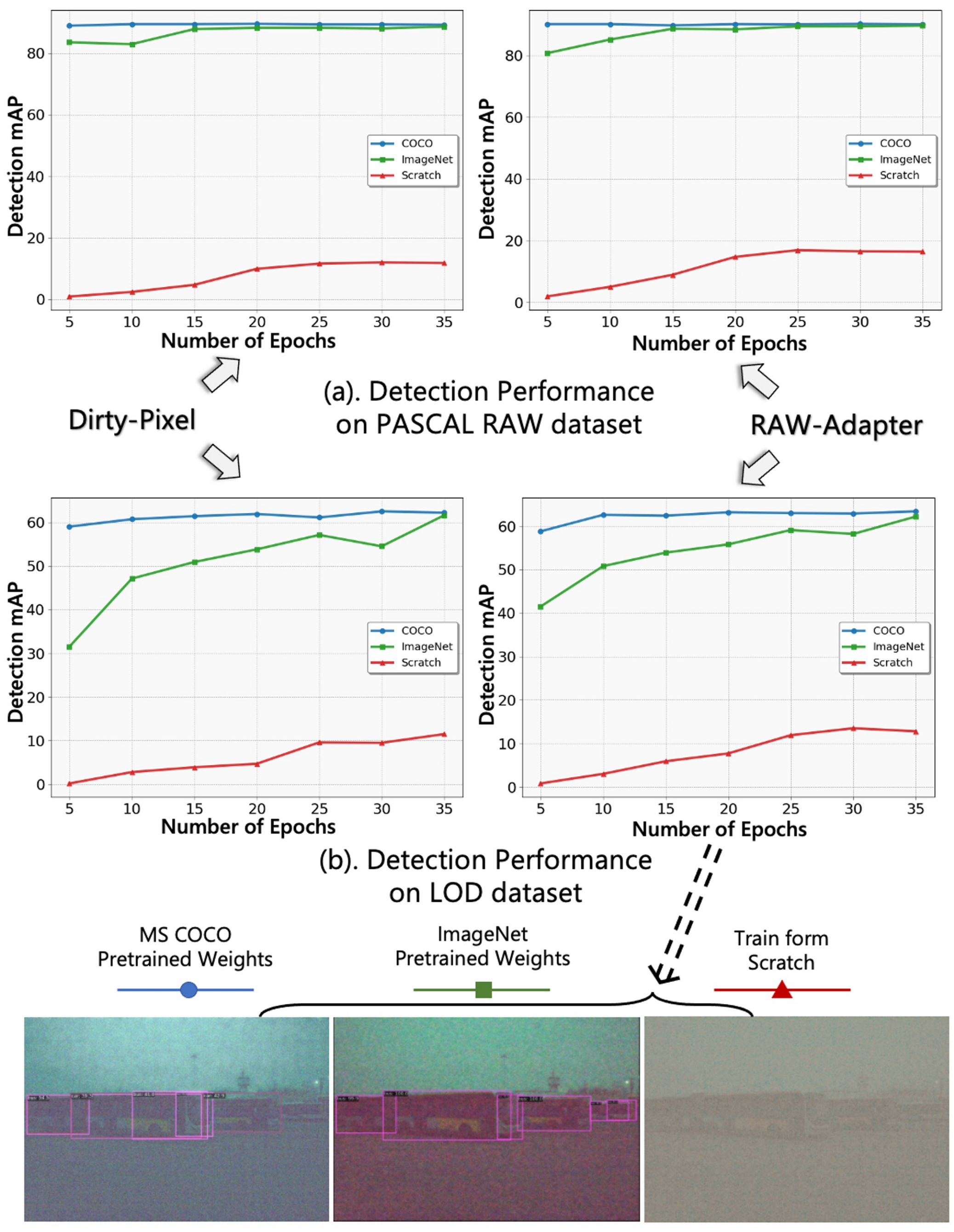}
    \vspace{-3mm}
    \caption{Performance of RAW-based object detection tasks with and without sRGB pre-trained weights. We analyze two methods: Dirty-Pixel~\cite{steven:dirtypixels2021} and our RAW-Adapter. Blue line represents trained with MS COCO~\cite{COCO_dataset} pre-train weights, the green line indicates ImageNet~\cite{imagenet_cvpr09} pre-train weights, and the red line signifies training from scratch. We separately present the detection performance on the PASCAL RAW~\cite{omid2014pascalraw} and LOD~\cite{LOD_BMVC2021} datasets.}
    \label{fig:pretrain_weights}
    \vspace{-5mm}
\end{figure}

\IEEEPARstart{I}{n} recent years, the rapid developments in camera sensor technology and computational photography have renewed interest in leveraging unprocessed camera RAW images for computer vision tasks. Unlike sRGB images—whose quality is affected by complex, lossy image signal processor (ISP) pipelines—RAW images preserve the scene’s linear radiometric data, full dynamic range, and intrinsic noise characteristics~\cite{Dancing_under_light,Wei_2020_CVPR,raw2raw}. This unadulterated data format provides a faithful scene representation and a reliable foundation in various sub-areas.
For example, image denoising techniques~\cite{brooks2019unprocessing,zamir2020cycleisp,li2025dualdn} exploit the inherent noise distributions in RAW to reduce sensor noise while preserving fine textures and edges. Additionally, novel view synthesis (NVS) frameworks~\cite{RAW_NeRF,jin2024le3d} leverage the rich information from RAW to generate more accurate and coherent 3D scene representations. Moreover, high-level visual algorithms (e.g., object detection and semantic segmentation) also benefit from RAW data, thus achieving higher accuracy under various challenging lighting conditions~\cite{Hardware_in_the_loop,ijcv_instance_dark,li2024towards}. These advancements underscore the substantial potential of camera RAW data across various applications. In autonomous driving, RAW images provide unprocessed, rich data for precise scene interpretation; in wildlife monitoring, they reveal subtle details even in extremely dark conditions; and in intelligent security, they ensure consistent reliability across challenging environments. Consequently, leveraging RAW images drives both technological innovation and the development of robust, high-performance real-world systems.

\begin{figure*}
    \centering
    \includegraphics[width=0.95\linewidth]{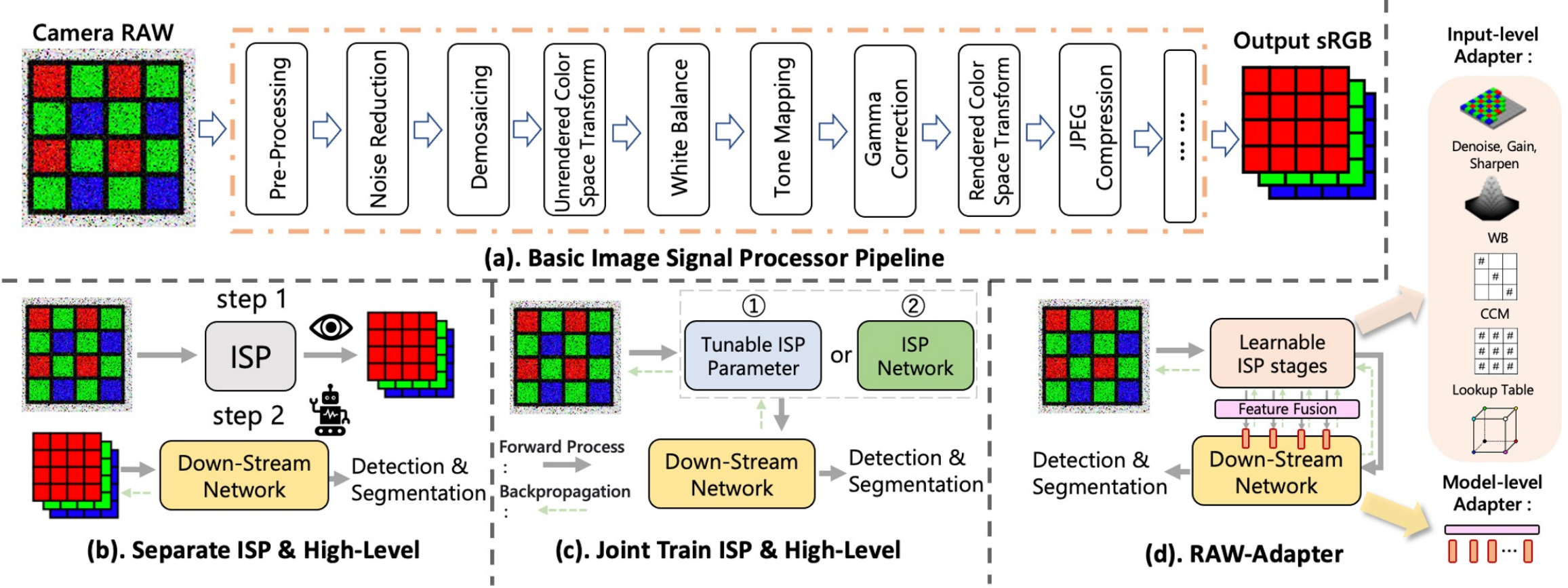}
    \vspace{-3mm}
    \caption{(a). An overview of basic image signal processor (ISP) pipeline. (The sequence or configuration may vary depending on the specific ISP settings of different manufacturers~\cite{Michael_eccv16,ISP_2005,Tseng2022NeuralPhotoFinishing}.) (b). ISP and current high-level vision models have different objectives. (c) Previous methods optimize ISP with down-stream high-level vision models. (d) Our proposed RAW-Adapter.}
    \label{fig:motivation}
    \vspace{-3mm}
\end{figure*}


Despite the promising prospects, significant challenges still remain. Most state-of-the-art (SOTA) vision models are pre-trained on large-scale sRGB datasets (e.g., ImageNet~\cite{imagenet_cvpr09}, MS-COCO~\cite{COCO_dataset}), creating a pronounced domain gap with camera RAW inputs. Training from scratch without these sRGB pre-trained weights can severely degrade performance (see Fig.~\ref{fig:pretrain_weights} for an example). Moreover, conventional ISP pipelines optimize RAW inputs for human perception rather than machine interpretation, often discarding or distorting subtle physical cues in RAW data\cite{ISP_2005,Michael_eccv16,heide2014flexisp,Camera_Net,Mobile_Computational}. As Fig.~\ref{fig:motivation}(a) illustrates, each ISP module targets specific perceptual goals instead of addressing following task requirements (see Fig.~\ref{fig:motivation}(b)), and the opaque nature of commercial ISPs (e.g., Adobe Camera RAW$^\circledR$) further complicates repurposing their processing for downstream algorithmic tasks.

To tackle these challenges, researchers have explored strategies that jointly optimize a tunable ISP and the subsequent computer vision network (see Fig.~\ref{fig:motivation}(c)). Broadly, these approaches fall into two categories. The first preserves the modular structure of the traditional ISP by designing differentiable modules~\cite{Hardware_in_the_loop,yu2021reconfigisp,wang2024adaptiveisp}, often leveraging optimization techniques such as CMA-ES~\cite{CMA-ES}. The second replaces the ISP entirely with a neural network module (e.g., UNet~\cite{UNet})~\cite{steven:dirtypixels2021,RAW_OD_CVPR2023,Attention_ISP_ECCV}, which typically introduces a significant computational burden for high-resolution inputs. However, both paradigms have largely treated the ISP and the vision network as independent components, thus neglecting the potential for a more synergistic integration between RAW inputs and sRGB pre-trained models.

To better bridge the gap between information-rich RAW data and knowledge-rich sRGB pre-trained models, we propose \textbf{RAW-Adapter}, a framework that redefines the relationship between camera RAW inputs and sRGB models. Instead of relying on complex ISP modules or deep neural network encoders, our approach simplifies the input stage while fostering a tighter integration with sRGB pre-trained models at the model level. Inspired by recent advances in prompt learning and adapter tuning~\cite{Prompt_NLP,VPT_ECCV2022,Vit_adapter,potlapalli2023promptir}, we develop two novel adapter approaches (input-level and model-level adapters) to enhance the connection between the two. As illustrated in Fig.~\ref{fig:motivation}(d). Input-level adapters transform RAW data into a suitable format for the backend network. Following traditional ISP design, we predict key ISP stage parameters to translate input RAW images for target downstream vision tasks. To ensure differentiability while maintaining a lightweight design, we incorporate Query Adaptive Learning (QAL) and Implicit Neural Representation (INR) into core ISP processes.
Model-level adapters utilize prior knowledge from input-level adapters by extracting intermediate ISP-stage features. These features are then embedded into the downstream network, allowing the model to incorporate ISP priors and jointly contribute to final machine-vision perception tasks.

\begin{figure*}[t]
    \centering
    \includegraphics[width=0.98\linewidth]{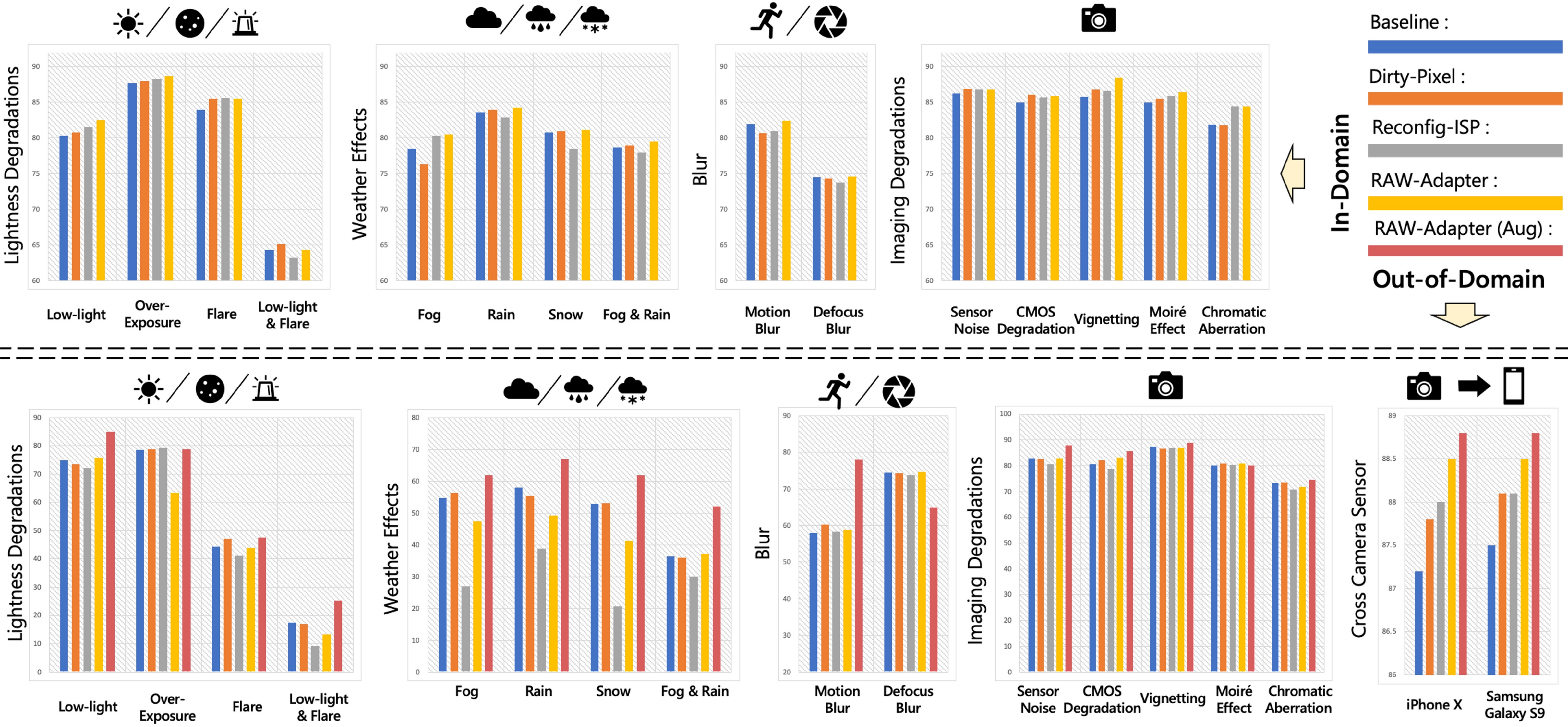}
    \vspace{-2mm}
    \caption{We propose \textbf{RAW-Bench} to evaluate the performance of current RAW-based vision frameworks against common corruptions, including lighting, weather, blur, camera imaging degradation, and variations in camera color response. Here, we present results on the PASCAL RAW-D dataset.}
    \label{fig:robustness_info}
    \vspace{-4mm}
\end{figure*}

Beyond the architectural improvements, practical applications also need considering model performance under diverse real-world corruptions~\cite{arnab2018robustness} (e.g., lighting, weather). Especially in camera RAW-based tasks, it is common to encounter inconsistencies between the sensor used for training data and that used for testing data (e.g., training with a Nikon D3200 DSLR RAW and testing with an iPhone X RAW). To this end, we propose \textbf{RAW-Bench}, a comprehensive benchmark incorporating 17 types of RAW-based common corruptions, thoroughly considering various lighting degradations, weather effects, blurriness, camera imaging degradations, and variations in camera color response (see Fig.~\ref{fig:robustness_info} and Fig.~\ref{fig:robustness}). To better evaluate the real-world generalization ability of RAW-Adapter, we conduct comparisons on \textbf{RAW-Bench} using different input image types (e.g., RAW, log-RGB), mainstream ISP algorithms, and other joint-training solutions to validate our performance. Additionally, we design experiments to separately evaluate the in-domain (ID) and out-of-domain (OOD) performance of the aforementioned methods.

Our contributions could be summarized as follows:

\begin{itemize}
   \item We introduce \textbf{RAW-Adapter}, a lightweight framework that enhances the integration of sRGB pre-trained models with camera RAW through adapter tuning. It addresses differences at both the input and model levels: input-level adapters efficient optimize key ISP parameters, while model-level adapters leverage this input-stage information to further improve downstream performance. Designed for both effectiveness and efficiency, RAW-Adapter minimizes computational overhead while enhances the performance of various downstream visual models.

    \item We introduce \textbf{RAW-Bench}, a benchmark encompassing 17 common corruptions that impact RAW-based visual performance. We  analyze their influence on existing methods across various factors, including lightness, weather conditions, blurriness, imaging degradations, and cross-sensor performance, offering valuable insights for future research. Furthermore, we propose a simple yet effective RAW-based data augmentation strategy to enhance RAW-Adapter’s performance and OOD robustness.
    

    \item Through comprehensive comparisons across different input data types, mainstream ISP solutions, and joint-training strategies in object detection and semantic segmentation tasks—evaluated under diverse corruption scenarios, we demonstrate that RAW-Adapter achieves state-of-the-art (SOTA) performance while exhibiting exceptional robustness against real-world corruptions.


    

\end{itemize}

\section{Related Works}
\label{sec2:related}

\subsection{Image Signal Processor (ISP)}

\subsubsection{Traditional ISP methods}

Typically, a camera's Image Signal Processor (ISP) pipeline is essential for rendering captured images from the camera's internal raw space into a standard display color space (e.g., sRGB). Traditional ISP pipelines are formulated as a series of manually crafted modules executed sequentially~\cite{ISP_2005,Michael_eccv16,Mobile_Computational,ICIP_Parameter_Tuning}. These modules include representative steps such as demosaicing, white balance, noise reduction, tone mapping, and color space transformation.
Many of these steps rely on real-world physical priors and human experience-based design approaches. For example, white balance in ISPs often depends on illuminant estimation~\cite{illumination_estimation_pami} as prior information or employs statistical hypotheses, such as the gray-world~\cite{gray_world} and white-patch~\cite{white_patch}. There also exist alternative designs of ISP processes such as Heide~\textit{et al.}~\cite{heide2014flexisp} designed FlexISP which combines numerous ISP blocks into a unified optimization block, and Hasinoff~\textit{et al.}~\cite{Siggraph16_low_light_ISP} modified some of the traditional ISP steps for burst photography under extreme low-light conditions. Traditional ISP methods are highly interpretable, support efficient troubleshooting and optimization, and do not require large-scale training datasets. However, they are also intrinsically tied to specific hardware, requiring expert manual tuning and consistently lacking end-to-end translation capability from RAW to sRGB~\cite{Mobile_Computational,SID}.

\subsubsection{Deep-learning based ISP methods}

With the advent of the deep learning era, many researchers have aimed to develop end-to-end pipelines for RAW-to-sRGB mapping~\cite{SID,hu_white_box,Deep_ISP,Camera_Net,PyNet,RAW-to-sRGB_ICCV2021,Meta_ISP,AAAI_learnable_ISP,jincvpr23dnf,Tseng2022NeuralPhotoFinishing,pan2024mambasci,He2024EnhancingRW}.
For example, Chen~\textit{et al.}\cite{SID} proposed SID, which replaces traditional ISP steps with a UNet\cite{UNet} to translate low-light RAW data into normal-light sRGB images. Similarly, Hu~\textit{et al.}\cite{hu_white_box} introduced a learnable white-box solution with 8 differentiable filters, allowing for the inspection of each step in the white-box process. Zhang~\textit{et al.}\cite{RAW-to-sRGB_ICCV2021} proposed LiteISP, which additionally addresses the impact of color inconsistency on image alignment. Tseng~\textit{et al.}~\cite{Tseng2022NeuralPhotoFinishing} introduced Neural Photo-Finishing, which not only completes ISP pipeline but also supports style transfer and adversarial photo editing.
However, a major drawback of deep network-based ISP models is their inherent reliance on the training dataset, which often limits their generalization performance.


\subsubsection{sRGB-to-RAW de-rendering}

Another line of research focuses on translating sRGB images back to the RAW domain~\cite{reverse_ISP_ICIP,64KB_RGB2RAW,Nam2017ModellingTS,Wang_2023_CVPR,brooks2019unprocessing,invertible_ISP,zamir2020cycleisp,kim2024paramisp}, mitigating challenges such as large storage requirements and limited support in many imaging applications. Approaches for converting RGB to RAW include recording camera metadata~\cite{64KB_RGB2RAW,Wang_2023_CVPR}, manually designed unprocessing pipelines~\cite{reverse_ISP_ICIP,brooks2019unprocessing}, and neural network-based methods~\cite{Nam2017ModellingTS,zamir2020cycleisp,invertible_ISP,kim2024paramisp}. Many of these methods support bidirectional conversions between sRGB and RAW. For example, CycleISP~\cite{zamir2020cycleisp} employs cascaded sRGB-to-RAW and RAW-to-sRGB networks, InvISP~\cite{invertible_ISP} uses a flow-based architecture for bidirectional mapping, and ParamISP~\cite{kim2024paramisp} leverages parameters from ParamNet to control the conversion direction. Beyond conversion between RAW and sRGB, ongoing research focuses on leveraging RAW images for downstream high-level computer vision tasks.

\vspace{-2mm}
\subsection{RAW-based High-level Vision Tasks}

\subsubsection{Architecture Improvements}

To effectively leverage the rich information in RAW data for downstream high-level vision tasks, early methods proposed performing vision algorithms directly on RAW data~\cite{buckler2017reconfiguring,Bayer_RAW_HOG}, bypassing the ISP process. However, these approaches fail to account for the camera noise introduced during the conversion from photons to RAW, especially under low-light conditions~\cite{Wei_2020_CVPR,Dancing_under_light,zhuoxiao_li}. Additionally, training from scratch on RAW data forfeits the benefits of large-scale pre-trained visual models on sRGB data (see Fig.~\ref{fig:pretrain_weights}). This challenge is further compounded by a significant disparity in dataset sizes: RAW image datasets, such as PASCAL RAW~\cite{omid2014pascalraw} (4,259 images) and LOD~\cite{LOD_BMVC2021} (2,230 images), are much smaller than current RGB datasets. For instance, ImageNet~\cite{imagenet_cvpr09} contains over 1 million images, while SAM~\cite{Segment_Anything} is trained on 11 million images.

Most subsequent research has focused on integrating the ISP with backend computer vision models~\cite{Hardware_in_the_loop,steven:dirtypixels2021,RAW_OD_CVPR2023,Vision_ISP_ICIP,yu2021reconfigisp,Attention_ISP_ECCV,DynamicISP_2023_ICCV,Guo_2024_CVPR,wang2024adaptiveisp}. For example, Wu~\textit{et al.}\cite{Vision_ISP_ICIP} first proposed VisionISP and emphasized the differences between human and machine vision, introducing several trainable modules in the ISP to enhance backend object detection performance. Since then, several methods have sought to replace traditional non-differentiable ISPs with differentiable ISP networks\cite{Hardware_in_the_loop,yu2021reconfigisp,wang2024adaptiveisp}. For instance, ReconfigISP employs a Neural Architecture Search (NAS) approach to identify the optimal ISP configuration. Alternatively, some methods directly replace the ISP process with encoder networks~\cite{steven:dirtypixels2021,Attention_ISP_ECCV,DynamicISP_2023_ICCV,RAW_OD_CVPR2023}. For example, Dirty-Pixel~\cite{steven:dirtypixels2021} utilizes a stack of UNets as a pre-encoder.
However, training two consecutive networks simultaneously imposes substantial computational burden, and previous research has rarely explored how to efficiently fine-tune current mainstream sRGB-based vision models for camera RAW data.


\subsubsection{RAW-based Vision Under Degradation} Meanwhile, research such as Rawgment~\cite{Rawgment_2023_CVPR} focuses on achieving realistic data augmentation directly on RAW images for robust perception. Some studies address the decline in computer vision performance caused by specific steps within the in-camera processing pipeline or ISP, such as white balance errors~\cite{Error_WB_Vision}, auto-exposure errors~\cite{Exposure_CVPR2021}, camera motion blur~\cite{Blurry_detection}, and undesirable camera noise~\cite{cui2022exploring}. Additionally, several RAW-based datasets have been proposed to account for specific real-world scenarios. For instance, LOD~\cite{LOD_BMVC2021} and ROD~\cite{RAW_OD_CVPR2023} consider low-light conditions. Very recently, AODRaw~\cite{li2024towards} has been introduced, further incorporating camera RAW data under diverse adverse weather conditions.

In this work, we propose \textbf{RAW-Bench}, which extends prior considerations of lighting and weather conditions by additionally incorporating flare, snow, various types of blurriness, camera imaging degradations, and cross-sensor variations. We unify these factors into a comprehensive benchmark consisting of 17 types of common corruptions, designed to evaluate models' in-domain (ID) and out-of-domain (OOD) performance.


\vspace{-1mm}
\subsection{Adapter Tuning in Computer Vision}

Adapters have become prevalent in the natural language processing (NLP) area, which introduce new modules in large language models (LLMs) such as~\cite{Prompt_NLP,stickland2019bert}. These modules enable task-specific fine-tuning, allowing pre-trained models to swiftly adapt to downstream NLP tasks. 

In the computer vision area, adapters have been adopted in various areas such as incremental learning~\cite{iscen2020memory} and domain adaptation~\cite{adapter_DA}. Recent years, a series of adapters have been proposed to investigate how to better utilize pre-trained visual models~\cite{Vit_adapter,VPT_ECCV2022,sam_hq,Adapter_point_cloud,Adapter_dense} or pre-trained vision-language models~\cite{sung2022vl,zhang2021tip}. These methods focus on utilizing prior knowledge to make pre-trained models quickly adapt to downstream tasks. Like ViT-Adapter~\cite{Vit_adapter} utilizes feature injector and feature extractor to adapt a pre-trained ViT~\cite{VIT} model to downstream detection and segmentation tasks, DAPT~\cite{Adapter_point_cloud} propose to adapt ViT model for point cloud analyze. Different like previous adapter research mostly focuses on RGB domain, our RAW-Adapter  further improves the alignment between camera RAW inputs and sRGB pre-trained visual models.

\begin{figure*}
    \centering
    \includegraphics[width=0.86\linewidth]{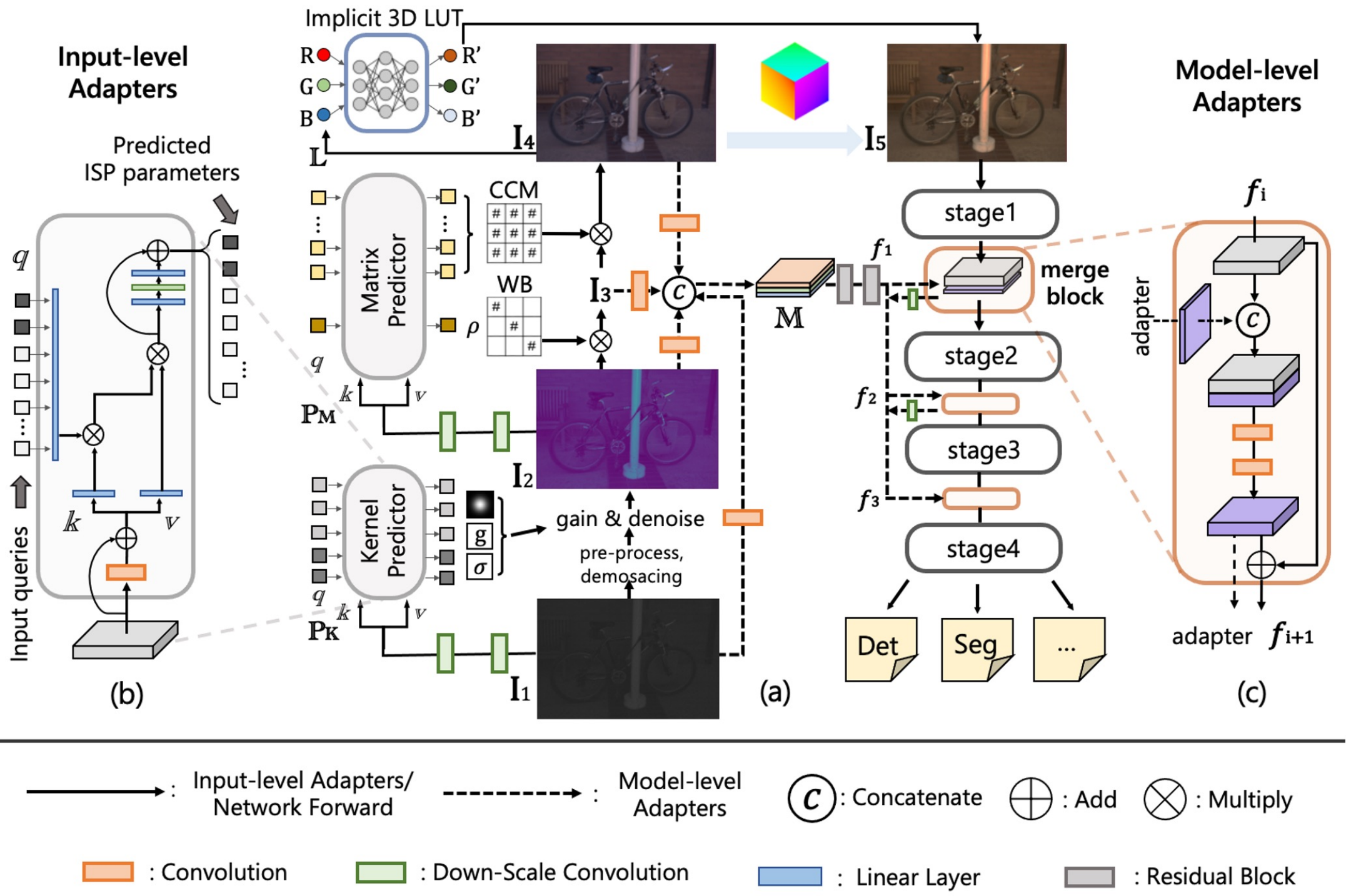}
    \vspace{-2mm}
    \caption{(a). Overall structure of RAW-Adapter. Solid line in left denotes input-level adapter's workflow (from input RAW data $\mathbf{I}_1$ $\rightarrow$ $\mathbf{I}_2$ $\rightarrow$ $\mathbf{I}_3$ $\rightarrow$ $\mathbf{I}_4$ $\rightarrow$ $\mathbf{I}_5$) and dotted line denotes model-level adapter's workflow, stage 1$\sim$4 means different stage of visual model backbone. (b). Detailed structure of kernel $\&$ matrix predictors $\mathbb{P_K}$, $\mathbb{P_M}$. (c). Detailed structure of model-level adapter $\mathbb{M}$'s merge block.}
    \label{fig:overview}
    \vspace{-2mm}
\end{figure*}

\section{RAW-Adapter}
\label{sec3:method}

In this section, we present our proposed \textbf{RAW-Adapter}. A brief overview of the conventional image signal processor (ISP) is provided in our supplementary materials. For the adapter design, we selectively omit certain steps to focus on key aspects of the camera ISP. Section~\ref{sec3.2input} details the input-level adapters, followed by an explanation of the model-level adapters in Section~\ref{sec3.3model}.

\subsection{Input-level Adapters}
\label{sec3.2input}

Fig.~\ref{fig:overview} provides an overview of RAW-Adapter. The solid lines in Fig.~\ref{fig:overview}(a) left represent input-level adapters. Input-level adapters are designed to convert the RAW image $\mathbf{I}_1$ into the machine-vision oriented image $\mathbf{I}_5$. This process involves digital gain $\&$ denoise, demosacing, white balance adjustment, camera color matrix, and color manipulation.

We maintain the ISP design while simultaneously using query adaptive learning (QAL) to estimate key parameters in the ISP stages. The QAL strategy is motivated by previous transformer models~\cite{DETR,cui2022you,cheng2021maskformer} and detailed structure is shown in Fig.~\ref{fig:overview}(b), input image $\mathbf{I}_{i\in(1,2)}$ would pass by 2 down-scale convolution blocks to generate feature, then feature pass by 2 linear layers to generate attention block's key $\mathbbm{k}$ and value $\mathbbm{v}$, while query $\mathbbm{q}$ is a set of learnable dynamic parameters, the ISP parameters would be predicted by self-attention calculation~\cite{vaswani2017attention}: 

\begin{equation}
    parameters = FFN(softmax(\frac{\mathbbm{q}  \cdot \mathbbm{k}^T}{\sqrt{d_k}}) \cdot \mathbbm{v}),
    \label{eq:predictor}
\end{equation}
where $FFN$ denotes the feed-forward network, includes 2 linear layers and 1 activation layer, the predicted parameters would keep the same length as query $\mathbbm{q}$. We defined 2 QAL blocks $\mathbb{P_K}$ and $\mathbb{P_M}$ to predict different part parameters.


The input RAW image $\mathbf{I}_1$ would first go through the pre-process  operations and a demosacing stage~\cite{Michael_eccv16,Mobile_Computational}, followed by subsequent ISP processes:

\subsubsection{Gain $\&$ Denoise}


Denoising algorithms always take various factors into account, such as input photon number, ISO gain level, exposure settings.   
Here we first use QAL block $\mathbb{P_K}$ to predict a gain ratio $g$~\cite{brooks2019unprocessing,SID} to adapt $\mathbf{I}_1$ in  different lighting scenarios, followed by an adaptive anisotropic Gaussian kernel to suppress noise under various noise conditions, $\mathbb{P_K}$ will predict the appropriate Gaussian kernel $k$ for denoising to improve downstream visual tasks' performance, the predicted key parameters are the Gaussian kernel's major axis $r_1$, minor axis $r_2$ (see Fig.~\ref{fig:para} left), here we set the kernel angle $\theta$ to $0$ for simplification.  Gaussian kernel $k$ at pixel $(x, y)$ would be: 


\begin{equation}
    k(x,y) = exp(-(b_0  x^2 + 2b_1  x y + b_2y^2)), 
\end{equation}
where: 
\begin{equation}
\begin{aligned}
    b_0 = \frac{cos(\theta)^2}{2{r_1}^2} & + \frac{sin(\theta)^2}{2{r_2}^2}, b_1 = \frac{sin(2\theta)}{4{r_1}^2}((\frac{r_1}{r_2})^2 - 1), \\
    & b_2 = \frac{sin(\theta)^2}{2{r_1}^2} + \frac{cos(\theta)^2}{2{r_2}^2}. 
\end{aligned}
\end{equation}

 After gain ratio $g$ and kernel $k$ process on image $\mathbf{I}_1$, $\mathbb{P_K}$ also predict a filter parameter $\sigma$ (initial at 0) to keep the sharpness and recover details of generated image $\mathbf{I}_2$. Eq.~\ref{eq:kernel_process} shows the translation from $\mathbf{I}_1$ to $\mathbf{I}_2$, where $\circledast$ denotes kernel convolution and filter parameter $\sigma$ is limited in a range of (0, 1) by a Sigmoid activation.  For more details please refer to our supplementary part.

\begin{equation}
\begin{aligned}
    & \mathbb{P_K}(\mathbf{I}_{1}, \mathbbm{q}) \rightarrow k\left\{r_1, r_2, \theta \right\}, g, \sigma,  \\
    & \mathbf{I}_2' = (g \cdot \mathbf{I}_1) \circledast k,  \\
    & \mathbf{I}_2 = \mathbf{I}_2' + (g \cdot \mathbf{I}_1 - \mathbf{I}_2') \cdot \sigma .
\label{eq:kernel_process}
\end{aligned}
\end{equation}





\subsubsection{White Balance $\&$ CCM Matrix}
\begin{figure}
    \centering
    \includegraphics[width=1.0\linewidth]{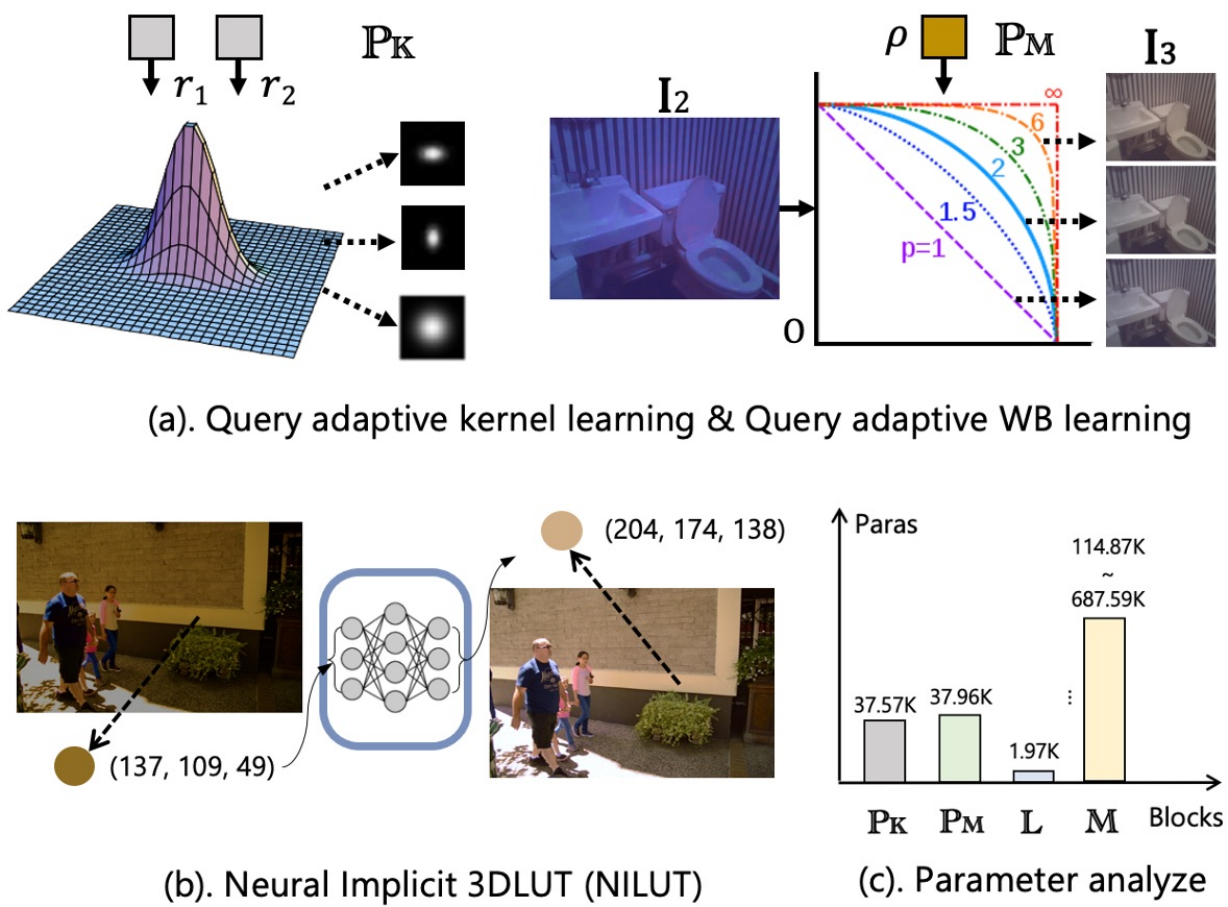}
    \vspace{-4mm}
    \caption{(a). We adopt query adaptive learning (QAL) to predict key parameters is ISP process. (b). An illustration of neural implicit 3DLUT (NILUT~\cite{conde2024nilut}) (c). We show RAW-Adapter different blocks' parameter.}
    \label{fig:para}
    \vspace{-3mm}
\end{figure}

White balance (WB) mimics the color constancy of the human vision system (HVS) by aligning ``white'' color with the white object, resulting in the captured image reflecting the combination of light color and material reflectance~\cite{ICCV_MAET,Error_WB_Vision,brooks2019unprocessing}. In our work, we hope to find an adaptive white balance for different images under various lighting scenarios. Motivated by the design of Shades of Gray (SoG)~\cite{Shades_of_gray} WB algorithm, where gray-world WB and Max-RGB WB can be regarded as a subcase, we've employed a learnable parameter $\rho$ to replace the gray-world's L-1 average with an adaptive Minkowski distance average (see Eq.~\ref{eq:matrix_process}), the QAL block $\mathbb{P_M}$ predicts Minkowski distance's hyper-parameter $\rho$, a demonstration of various $\rho$  is shown in  Fig.~\ref{fig:para}. After finding the suitable $\rho$, $\mathbf{I}_2$ would multiply to white balance matrix to generate $\mathbf{I}_3$.



Color Conversion Matrix (CCM) within the ISP is constrained by the specific camera model. Here we standardize the CCM as a single learnable 3$\times$3 matrix, initialized as a unit diagonal matrix $\mathbf{E}_3$. The QAL block $\mathbb{P_M}$ predicts 9 parameters here, which are then added to $\mathbf{E}_3$ to form the final 3$\times$3 matrix $\mathbf{E}_{ccm}$. Then $\mathbf{I}_3$ would multiply to CCM $\mathbf{E}_{ccm}$ to generate $\mathbf{I}_4$, equation as follow:

\begin{equation}
\begin{aligned}
    &  \mathbb{P_M}(\mathbf{I}_{2}, \mathbbm{q}) \rightarrow \rho, \mathbf{E}_{ccm},\\
    & m_{i \in (r,g,b)} = \sqrt[\rho]{avg(\mathbf{I}_{2}(i)^\rho)} /  \sqrt[\rho]{avg((\mathbf{I}_{2})^\rho)}, \\
    & \mathbf{I}_{3} = \mathbf{I}_{2} * \begin{bmatrix}
   m_r &  &  \\
   & m_g &  \\
    &  & m_b
  \end{bmatrix}, \quad  \mathbf{I}_{4} = \mathbf{I}_{3} * \mathbf{E}_{ccm}.
\label{eq:matrix_process}
\end{aligned}
\end{equation}


\subsubsection{Color Manipulation Process} In ISP, color manipulation process is commonly achieved through lookup table (LUT), such as 1D and 3D LUTs~\cite{Mobile_Computational}. Here, we consolidate color manipulation operations into a single 3D LUT, adjusting the color of image $\mathbf{I}_4$ to produce output image $\mathbf{I}_5$. Leveraging advancements in LUT techniques, we choose the latest neural implicit 3D LUT (NILUT)~\cite{conde2024nilut}, for its speed efficiency and ability to facilitate end-to-end learning in our pipeline~\footnote{To save memory occupy, we change channel number 128 in~\cite{conde2024nilut} to 32.}. We denote NILUT~\cite{conde2024nilut} as $\mathbb{L}$, $\mathbb{L}$ maps the input pixel intensities $R, G, B$ to a continuous coordinate space, followed by the utilization of implicit neural representation (INR)~\cite{sitzmann2019siren}, which involves using a multi-layer perceptron (MLP) network to map the output pixel intensities to $R', G', B'$:

\begin{equation}
    \mathbf{I}_5(R', G', B') =  \mathbb{L}(\mathbf{I}_4(R, G, B)).
\end{equation}

Image $\mathbf{I}_5$ obtained through image-level adapters will be forwarded to the downstream network's backbone. Furthermore, $\mathbf{I}_5$'s features obtained in backbone will be fused with model-level adapters, which we will discuss in detail.



\subsection{Model-level Adapters}
\label{sec3.3model}

Input-level adapters guarantee the production of machine vision-oriented image $\mathbf{I}_5$ for high-level models. However, information at the ISP stages ($\mathbf{I}_1 \sim \mathbf{I}_4$) is almost overlooked. Inspired by adapter design in current NLP and CV area~\cite{Prompt_NLP,Vit_adapter,VPT_ECCV2022}, we employ the prior information from the ISP stage as model-level adapters to guide subsequent models' perception. Additionally, model-level adapters promote a tighter integration between downstream tasks and the ISP stage. Dotted lines in Fig.~\ref{fig:overview}(a) represent model-level adapters.

Model-level adapters $\mathbb{M}$ integrate the information from ISP stages to the network backbone. As shown in Fig.~\ref{fig:overview}(a), we denote the stage $1 \sim 4$ as different stages in network backbone, image $\mathbf{I}_5$ would pass by backbone stages $1 \sim 4$ and then followed by detection or segmentation head. We utilize convolution layer $c_i$ to extract features from $\mathbf{I}_1$ through $\mathbf{I}_4$, these extracted features are concatenated as $\mathbb{C}(\mathbf{I}_{1\sim4}) = \mathbb{C}(c_1(\mathbf{I}_1), c_2(\mathbf{I}_2), c_3(\mathbf{I}_3), c_4(\mathbf{I}_4))$. Subsequently, $\mathbb{C}(\mathbf{I}_{1\sim4})$ go through two residual blocks~\cite{he2016resnet} and generate adapter $f_1$, $f_1$ would merge with stage $1$'s feature by a merge block, detail structure of merge block is shown in Fig.~\ref{fig:overview}(c), which roughly includes concatenate process and a residual connect, finally merge block would output network feature for stage $2$ and an additional adapter $f_2$. Then process would repeat in stage $2$ and stage $3$ of network backbone. We collectively refer to all structures related to model-level adapters as $\mathbb{M}$.


We show RAW-Adapter's different part parameter number in Fig.~\ref{fig:para}(c), including input-level adapters $\{ \mathbb{P_K}$ (37.57K), $\mathbb{P_M}$ (37.96K), $\mathbb{L}$ (1.97K)$\}$ and model level adapters $\mathbb{M}$ (114.87K $\sim$ 687.59K), model level adapters' parameter number depend on following backbone network. Total parameter number of RAW-Adapter is around 0.2M to 0.8M, much smaller than following backbones ($\sim$25.6M in ResNet-50~\cite{he2016resnet}, $\sim$197M in Swin-L~\cite{liu2021Swin}), also smaller than previous SOTA methods like SID~\cite{SID} (11.99M) and Dirty-Pixel~\cite{steven:dirtypixels2021} (4.28M).

After introducing our RAW-Adapter model, in Section~\ref{sec4:corruptions}, we would present the proposed \textbf{RAW-Bench}, which includes 17 types of common degradations and would be used to evaluate the robustness of RAW-Adapter and other methods~\cite{log-RGB,Colorimetric,Michael_eccv16,RAW-to-sRGB_ICCV2021,invertible_ISP,steven:dirtypixels2021,yu2021reconfigisp}. In Section~\ref{sec5:augmentation}, we will introduce the RAW-based data augmentation, which is designed to further enhance the performance of RAW-Adapter and improve its out-of-domain (OOD) generalization capability.









\section{RAW-Bench with RAW-based Common Corruptions}
\label{sec4:corruptions}

\begin{figure*}
    \centering
    \includegraphics[width=0.98\linewidth]{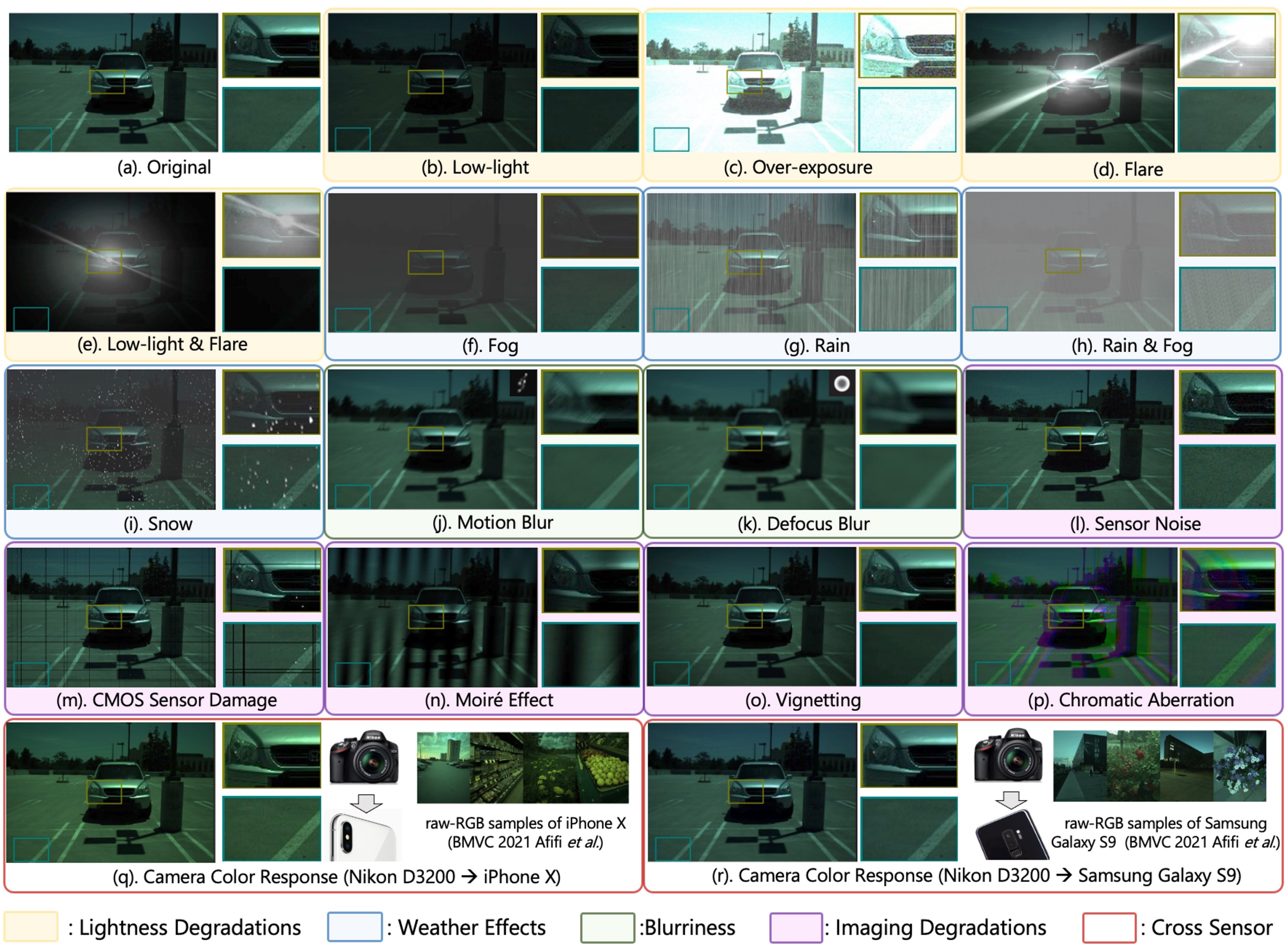}
    \vspace{-3mm}
    \caption{An ``car'' photo example from PASCAL RAW dataset~\cite{omid2014pascalraw}, which is better visualized by averaging the green channels and applying a gamma correction with an encoding gamma of 1/1.4. We synthesis various corruptions that may arise at different stages of camera imaging, such as lighting, weather, blur and inherent camera imaging degradations (e.g., sensor noise, CMOS sensor demage), as well as different camera color response functions (cross-sensor).}
    \label{fig:robustness}
    \vspace{-3mm}
\end{figure*}

In this section, we propose \textbf{RAW-Bench}, which includes various common corruptions in camera RAW space. We selected two datasets, PASCAL RAW~\cite{omid2014pascalraw} and iPhone XS Max~\cite{RAW_segment_dataset}, both captured under relatively normal conditions to synthesize corruptions. The PASCAL RAW dataset was transformed into PASCAL RAW-D as a robustness benchmark for object detection, while the iPhone XS Max dataset was transformed into iPhone XS Max-D for semantic segmentation. We didn't use datasets like~\cite{LOD_BMVC2021,RAW_OD_CVPR2023,li2024towards} as they were already captured under diverse lighting $\&$ weather conditions. Next, we introduce common corruptions that may occur in the camera RAW space, such as lightness degradations, weather effects, blurriness, imaging degradations, and variations in camera color response functions, and describe the corresponding degradation synthesis pipeline (see Fig.~\ref{fig:robustness} for a visualization). It is worth noting that previous vision robustness benchmarks~\cite{iclr_cls_robustness,kar20223d} are mostly built on rendered sRGB data, whereas our work focuses on the degradations occurring in the camera RAW space. Accordingly, we consider only corruptions that may occur in RAW and do not address degradations related to the ISP stage, such as those caused by incorrect white balance settings\cite{Error_WB_Vision} or ghosting artifacts introduced by chroma denoising~\cite{Siggraph16_low_light_ISP}.



\vspace{-2mm}
\subsection{Lightness Degradations}
\textbf{Low-light:} Both natural dark environments and camera exposure settings (ISO, f-number, exposure time) influence the brightness of captured RAW data. Since light intensity and environmental irradiance have a linear relationship with RAW data, we adopted the synthesis method from our conference version~\cite{raw_adapter} and previous works~\cite{ICCV_MAET,Li_2024_CVPR} to generate low-light RAW data, as follow:
\begin{equation}
\begin{aligned}
    &x_{n} \sim N(\mu = lx, \sigma^{2} = \delta_{r}^2 + \delta_{s}lx)\\
    &y = lx + x_{n},
\end{aligned}
\label{eq:low-light}
\end{equation}
where $x$ denotes input RAW data and $y$ denotes degraded RAW data, $\delta_{s} = \sqrt{S}$ denotes shot noise while $\sqrt{S}$ is signal of the sensor, $\delta_{r}$ denotes read noise, $l$ denotes the 
light intensity parameter which randomly chosen from [0.05, 0.4] to synthesis low-light corruption. An example is shown in Fig.~\ref{fig:robustness}(b).


\textbf{Overexposure:} Camera exposure errors and strong natural light can also significantly impact vision tasks~\cite{Exposure_CVPR2021,raw_adapter}. For overexposed RAW synthesis, we applied the same synthesis formula (Eq.\ref{eq:low-light}) used for low-light images, with the light intensity parameter $l$ randomly selected from the range [3.5, 5.0]. An example is shown in Fig.~\ref{fig:robustness}(c).


\textbf{Flare:} 
Photographs of scenes with strong light sources often exhibit lens flare, with its patterns influenced by lens optics, light source position, manufacturing imperfections, and accumulated scratches or dust~\cite{flare_remove_iccv,dai2022flare7k}. The local overexposure and artifacts caused by lens flare can negatively impact downstream tasks. Following~\cite{flare_remove_iccv}, we synthesize flare-corrupted images as follows:
\begin{equation}
    y = x + F + N(0, \sigma^2).
    \label{eq:flare}
\end{equation}
Here, $x$ denotes the flare-free input RAW data, $F$ represents the flare-only image, where we randomly choose 10 patterns from work~\cite{dai2022flare7k}, and $N(0, \sigma^2)$ denotes Gaussian noise with variance sampled per image from a scaled chi-square distribution, $\sigma^2 \sim \mathcal{X}^2$. An example is shown in Fig.~\ref{fig:robustness}(d).

\textbf{Low-light $\&$ Flare (Low $\&$ Flare):} 
Artificial lights in nighttime environments (such as car headlights and neon lights) can have an even greater negative impact~\cite{dai2022flare7k,dai2023flare7kpp}. Therefore, we further incorporate flare degradation into low-light conditions (Eq.~\ref{eq:low-light}), as follows:
\begin{equation}
    y = lx + F + x_{n},
\end{equation}
the light intensity parameter $l$ and noise $x_{n}$ are same as Eq.~\ref{eq:low-light}, an example is shown in Fig.~\ref{fig:robustness}(e).

\subsection{Weather Effects}

\textbf{Fog:} Foggy or hazy environments have a significant impact on computer vision tasks~\cite{kar20223d}. Following Koschmieder's Law, we synthesize the foggy effect on camera RAW data as follows:
\begin{equation}
    y = x\cdot t(x) +  \mathcal A \cdot (1-t(x)),
    \label{eq:fog}
\end{equation}
where $x$ denotes original RAW data and $y$ denotes the foggy degraded RAW data, $\mathcal A$ represents the global atmospheric light, and $t(x) = e^{-\beta d(x)}$ is medium transmission map. Here, $\beta$ denotes the atmosphere scattering parameter, and $d(x)$ stands for the depth of the scene $x$. We use the offline depth estimation method Depth-Anything model~\cite{depthanything} to predict depth  $d(x)$, then set hyper-parameters $\mathcal A$ and $\beta$ same as work~\cite{kar20223d}. An example is shown in Fig.~\ref{fig:robustness}(f).

\textbf{Rain:} Rain environments also has a significant impact on both human and machine vision systems~\cite{Rain_vision}. Here we take imgaug toolbox~\cite{imgaug} to add rain streak effect on camera RAW data, to synthesis rainy RAW data $y$ as follows: 
\begin{equation}
    y = x + \sum_{i=1}^{N} \mathcal R_i,
    \label{eq:rain}
\end{equation}
where x is rain-free input RAW data, $\mathcal R_i$ denotes the added rain layer. An example is shown in Fig.~\ref{fig:robustness}(g).

\textbf{Rain $\&$ Fog:} When considering heavy rain, it is essential to account for the combined effects of rain streaks and fog~\cite{valanarasu2022transweather}. Following~\cite{valanarasu2022transweather}, here we synthesize the rain $\&$ fog effect as follows:
\begin{equation}
    y = (x + \sum_{i=1}^{N} R_i)\cdot t(x) + \mathcal A \cdot (1-t(x)),
    \label{eq:rain_fog}
\end{equation}
the equation is a combination of Eq.~\ref{eq:fog} and Eq.~\ref{eq:rain}, an example is shown in Fig.~\ref{fig:robustness}(h).

\textbf{Snow:} Snow is also an important factor in adverse weather conditions, we follow previous work~\cite{valanarasu2022transweather} to complete the synthesis of snow, the equation is as follows:
\begin{equation}
    y = x \cdot (1-z) + z \cdot \mathcal S 
\end{equation}
where $z$ denotes the snow mask region, $\mathcal S$ denotes the snow flakes ($\mathcal S$ pattern taken from imgaug toolbox~\cite{imgaug}). An example is shown in Fig.~\ref{fig:robustness}(i).


\subsection{Blurriness}

\textbf{Motion Blur:} Motion blur occurs due to relative motion between the camera and the scene during exposure, leading to a loss of sharpness where objects appear smeared or stretched along the motion direction~\cite{PAMI_motion_blur}. To synthesize motion blur on demosaiced camera RAW, we use the point spread function (PSF) from the imgaug toolbox~\cite{imgaug}. An example is shown in Fig.~\ref{fig:robustness}(j).

\textbf{Defocus Blur:} Defocus blur occurs when a wide aperture prevents light rays from converging properly, causing scene points outside the camera’s focus distance to appear blurred~\cite{defocus_blur_pami}. To synthesize defocus blur on demosaiced camera RAW, we use the point spread function (PSF) from the imgaug toolbox~\cite{imgaug}. An example is shown in Fig.~\ref{fig:robustness}(k).


\subsection{Camera Imaging Degradations}

\textbf{Sensor Noise:} Noise in camera RAW images originates from various stages in the imaging pipeline, spanning the conversion processes from photons to electrons, then to voltage, and finally to digital numbers~\cite{brooks2019unprocessing,Wei_2020_CVPR}. In this work, we follow the modeling approach in~\cite{brooks2019unprocessing,ICCV_MAET}, which combines shot noise (a photon-related random process) and read noise (introduced during electronic readout) into a unified representation, and additionally consider analog-to-digital converter (ADC) quantisation noise, equation as follows:
\begin{equation}
\begin{aligned}
    &x_{n} \sim N(\mu = x, \sigma^{2} = \delta_{r}^2 + \delta_{s}x)\\
    & x_{quan} \sim U(-\frac{1}{2B}, \frac{1}{2B})\\
    &y = x + x_{n} + x_{quan},
\end{aligned}
\label{eq:noise}
\end{equation}
here, shot-read noise setting $x_{n}$ is the same as Eq.~\ref{eq:low-light}, where $x_{quan}$ denotes the quantisation noise related to $B$ bits, which follows a uniform distribution, and $B$ is set to 12 following the bits number in the PASCAL RAW dataset~\cite{omid2014pascalraw} and iPhone XS Max-D dataset~\cite{RAW_segment_dataset}. An example is shown in Fig.~\ref{fig:robustness}(l).

\textbf{CMOS Sensor Damage (CMOS-SD):} Aging CMOS sensors can cause black lines due to pixel row failures or readout circuit issues, and white dots from hot pixels, increased thermal noise, or fixed pattern noise. We model this corruption by introducing black line and white dot  damage into the digital signal, simulated based on the laser spot model~\cite{CMOS_deg}. An example is shown in Fig.~\ref{fig:robustness}(m).

\textbf{Moiré Effect:} Moiré patterns appear when the sampling frequency of the camera is insufficient to capture high-frequency details, leading to aliasing as described by the Nyquist Sampling Theorem~\cite{Amidror2000TheTO}. To synthesize Moiré patterns on camera RAW images, we create a stripes pattern~\cite{yue2023recaptured} and overlay it onto the RAW image, adjust the pattern’s frequency, angle, and transparency, and then blend the layers to produce the Moiré effect. An example is shown in Fig.~\ref{fig:robustness}(n).

\textbf{Vignetting:} Vignetting refers to the decrease in brightness or saturation at the edges of an image compared to the center. Several factors can contribute to vignetting, like the use of thick or stacked filters and improper lens hoods~\cite{Vignetting,Vignetting_PAMI}. Here, we follow the work~\cite{Vignetting} using 2D Gaussian vignette
filter to perform vignetting effects on camera RAW images, an example is shown in Fig.~\ref{fig:robustness}(o).

\textbf{Chromatic Aberration (CA):} Chromatic aberration occurs when the camera lens fails to focus all wavelengths of light at the same point, resulting in color ``fringes'' along the boundaries between dark and bright areas of the image~\cite{Chromatic_Aberration,Chromatic_syn}. Following~\cite{Chromatic_syn} we synthesize the chromatic aberration effect using radial distortions (k1, second order), which is implemented as a per-channel (red, green, blue) variation of the k1 radial distortion, an example is shown in Fig.~\ref{fig:robustness}(p).

\subsection{Camera Color Response Functions (Cross Sensor)}


In addition to common corruptions in camera RAW space, raw-RGB colors vary due to differences in spectral sensitivity across sensor makes~\cite{raw2raw,raw2raw_semi,rawformer,raw2raw_acmmm}. Mathematically, the raw-RGB color $x$ can be described as follows~\cite{raw2raw,raw2raw_semi}:
\begin{equation}
    x = \int_{\omega} \rho(\lambda) \cdot R(\lambda) \cdot L(\lambda) \ d\lambda.
\label{eq:raw2raw}
\end{equation}
here $\lambda$ represents the wavelength, $\omega$ denotes the visible spectrum (380 - 720 nm), $\rho$ denotes spectral power distribution, and $R$ denotes captured
object spectral reflectance properties, while $L$ denotes the camera sensor sensitivity. The differences in camera sensor sensitivities can lead to variations in the final captured RAW data. In our \textbf{RAW Bench}, we additionally treat the different camera color response functions $L$ as a type of common corruption to evaluate the model's OOD performances across different sensors. Take advantage of current SOTA unsupervised RAW-to-RAW translation models~\cite{rawformer,raw2raw_acmmm}, the input RAW data $x$ (captured by camera $A$) can be translated into target RAW data $y$ (same scene captured by camera $B$) without requiring paired training.
\begin{equation}
    y = f(x),
\label{eq:raw2raw_2}
\end{equation}
here $f(\cdot)$ denotes the sensor-to-sensor mapping. We use the translation method from~\cite{raw2raw_acmmm} to transform RAW data captured by a source sensor (e.g., Nikon D3200 DSLR~\cite{omid2014pascalraw}) to closely resemble RAW data captured by target sensors (e.g., iPhone X and Samsung Galaxy S9~\cite{raw2raw_semi}). Examples are shown in Fig.~\ref{fig:robustness}(q) and Fig.~\ref{fig:robustness}(r).




\section{RAW-based Data Augmentation}
\label{sec5:augmentation}

\begin{figure}
    \centering
    \includegraphics[width=1.0\linewidth]{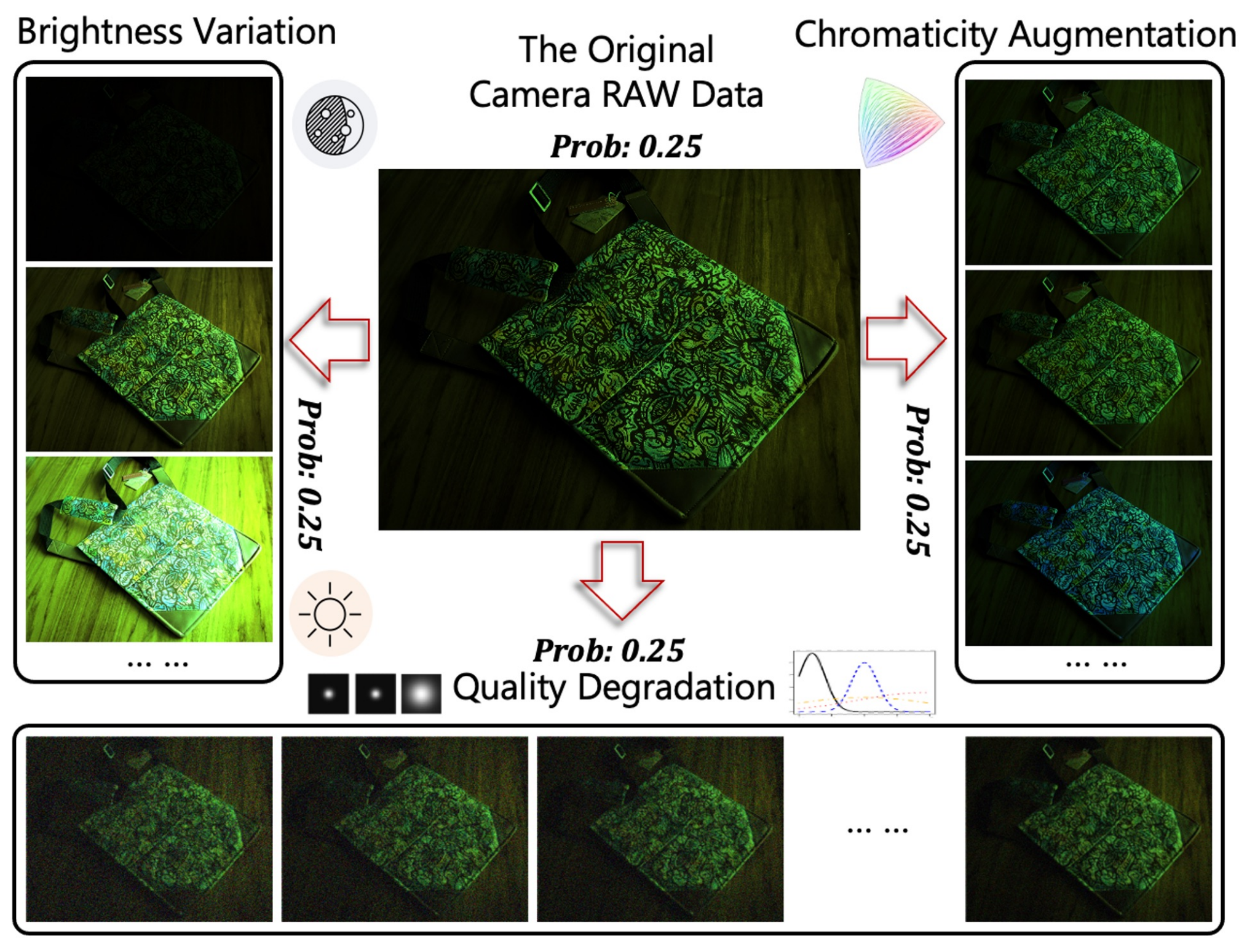}
    \vspace{-7mm}
    \caption{Overview of RAW-based data augmentation, which includes brightness variation, chromaticity augmentation and quality degradation.}
    \label{fig:augmentation}
    \vspace{-3mm}
\end{figure}

In this section, we introduce our RAW-based data augmentation solution to further improve the performance of RAW-Adapter and enhance its out-of-domain (OOD) robustness under various corruptions. Previous research has also shown the positive impact of data augmentation on improving model robustness~\cite{Aug_robustness_1,Aug_robustness_2}. Our augmentation method directly operates in the camera RAW domain and incorporates three key components: brightness variation, chromaticity augmentation, and quality degradation. An overview of the proposed method is shown in Fig.~\ref{fig:augmentation}. The detailed design of each augmentation component is described as follows:

\textit{(a). Brightness Variation:}  For the brightness variation, we map the input RAW to a random brightness level during training. Specifically, the input RAW image is scaled by a lighting coefficient $\omega$, where $\omega$ follows the distribution as:
\begin{equation}
    \omega \in \left\{
\begin{array}{ll}
\mathcal{TN}(0.2, 0.08; 0.01, 1.0), \quad prob = 0.5
\\
\mathcal{TN}(3.5, 1.0; 1.0, 5.0), \quad prob = 0.5
\end{array}
\right.
\end{equation}
where the $\mathcal{TN}(\mu, \sigma; min, max)$ denotes a truncated Gaussian distribution, $\mu$ is the mean, 
$\sigma$ is the standard deviation, and 
$min$ and $max$ represent the lower and upper bounds of the range, respectively. This ensures that the RAW images cover a diverse range of lightness levels during training.

\textit{(b). Chromaticity Augmentation:} Beyond brightness variation, we further incorporate chromaticity augmentation on the demosaicked 3-channel linear Raw-RGB data by multiplying each channel with random value, while ensuring that the sum of the three random values equals 3. Equation as follows: 
\begin{equation}
\begin{aligned}
    R' &= \omega_r \cdot R, G' = \omega_g \cdot G, B' = \omega_b \cdot B \\
    \omega_r&+ \omega_g + \omega_b = 3, \omega_r \in U(0.9, 1.1), \omega_b \in U(0.9, 1.1)
\end{aligned}
\end{equation}
A visualization of the augmented RAW data is in Fig.~\ref{fig:augmentation} right. We aim to maximize chromaticity color jitter augmentation while minimizing changes to the overall brightness.


\begin{table}[t]
\caption{A list of dataset and framework setting in our experiments.}
\vspace{-2mm}
\label{tab:dataset}
\centering
\renewcommand\arraystretch{1.9}

\begin{adjustbox}{max width = \linewidth}
\LARGE
\begin{tabular}{c|ccc}
\hline \hline 
\multicolumn{4}{c}{\fontsize{19}{20} \textit{\textbf{List of Dataset and Framework Setting}}}                                                                     \\ \hline
& \multicolumn{1}{c|}{\begin{tabular}[c]{@{}c@{}}PASCAL RAW~\cite{omid2014pascalraw}\\ (common corruptions)\end{tabular}} & \multicolumn{1}{c|}{LOD~\cite{LOD_BMVC2021}}                  & \begin{tabular}[c]{@{}c@{}}iPhone XS Max-D~\cite{RAW_segment_dataset}\\ (common corruptions)\end{tabular} \\ \hline
Task                                                        & \multicolumn{2}{c|}{Object Detection}                                                                                                       & Semantic Segmentation                                                           \\ \hline
Number                                                      & \multicolumn{1}{c|}{4259}                                                                       & \multicolumn{1}{c|}{2230}                 & 1153                                                                            \\ \hline
Seosor                                                      & \multicolumn{1}{c|}{Nikon D3200 DSLR}                                                           & \multicolumn{1}{c|}{Canon EOS 5D Mark IV} & iPhone XS Max                                                                   \\ \hline
Framework                                                   & \multicolumn{2}{c|}{RetinaNet~\cite{lin2017focal_loss} \& Sparse-RCNN~\cite{sparse_rcnn}}                                                                                               & SegFormer~\cite{xie2021segformer}       \\ \hline
Backbone                                                    & \multicolumn{2}{c|}{ResNet~\cite{he2016resnet}}    & MIT~\cite{xie2021segformer}        \\ \hline
Pre-train & \multicolumn{3}{c}{ImageNet~\cite{imagenet_cvpr09} pre-train weights}   \\ \hline \hline
\end{tabular}
\end{adjustbox}
\vspace{-2mm}
\end{table}

\textit{(c). Quality Degradation:} Motivated by previous works~\cite{AERIS_Aug, Rawgment_2023_CVPR}, here we further incorporate noise and blur as data augmentation for RAW data. Specifically, we adopt the degradation synthesis method from~\cite{AERIS_Aug}. 
For blur, we use both isotropic $k_{\text{iso}}$ and anisotropic $k_{\text{aniso}}$ Gaussian kernels; kernel size is uniformly sampled from  
$\left\{7\times7, 9\times9, ..., 21\times21 \right\}$.
For $k_{\text{iso}}$, the kernel width is uniformly drawn from $(0.1, 2.4)$. For $k_{\text{aniso}}$, the kernel angle is uniformly sampled from $(0, \pi)$, while the longer axis width is drawn from $(0.5, 6)$. For noise, we adopt a zero-mean additive white Gaussian noise (AWGN) model $n \sim N(0, \sigma)$, the variance $\sigma$ is randomly chosen from a uniform distribution $U(0, (0.1\cdot2^B)/2^B)$ and B denotes RAW's bits number. The aliasing for noise $\&$ blur follows the same approach as in~\cite{AERIS_Aug}.


The augmentations (a), (b), and (c) are applied alongside the original RAW images, with each being randomly selected during training with a probability of 0.25. Experiments show that our simple RAW-based data augmentation method significantly enhances the performance of RAW-Adapter and other algorithms, while also yielding substantial improvements in OOD robustness (see Fig.~\ref{fig:robustness_info}).

\begin{table}[t]
\caption{A list of comparison methods in our experiments.}
\vspace{-2mm}
\label{tab:compare_methods}
\renewcommand\arraystretch{1.5}
\Large
\resizebox{\linewidth}{!}{
\begin{tabular}{cccc}
\hline
\hline
\multicolumn{4}{c}{\textit{\textbf{List of Comparison Methods}}}                                                                                                    \\ \hline
\multicolumn{1}{c|}{Input data type}                                                       & RAW (Demosaiced) & log-RGB~\cite{log-RGB}         & colorimetric~\cite{Colorimetric} \\ \hline
\multicolumn{1}{c|}{\multirow{2}{*}{ISP methods}}                                     & Default ISP      & Karaimer \textit{et al.}~\cite{Michael_eccv16} & Lite-ISP~\cite{RAW-to-sRGB_ICCV2021}  \\ \cline{2-4} 
\multicolumn{1}{c|}{}                                                                 & InvISP~\cite{invertible_ISP}           & SID~\cite{SID}             & DNF~\cite{jincvpr23dnf}          \\ \hline
\multicolumn{1}{c|}{\begin{tabular}[c]{@{}c@{}}Joint-training\\ methods\end{tabular}} & Dirty-Pixel~\cite{steven:dirtypixels2021}      & Reconfig-ISP~\cite{yu2021reconfigisp}    & RAW-Adapter  \\ \hline\hline
\end{tabular}}
\label{tab:comparision}
\vspace{-2mm}
\end{table}

\vspace{-1mm}
\section{Experiments}
\label{sec6:exp}


In this section, we present our experiments. We begin with the datasets and experimental settings (Sec.\ref{sec6-1:setting}), followed by object detection (Sec.\ref{sec6-2:detection}) and semantic segmentation (Sec.\ref{sec6-3:seg}) results under the in-domain (ID) training setting. Finally, we report out-of-domain (OOD) generalization results (Sec.\ref{sec6-4:OOD}). Additional ablation studies and visualizations are provided in the supplementary material.

\begin{table*}[]
\caption{Comparison results (mAP$\%$) on PASCAL RAW-D dataset. Red background color shows
best mAP result meanwhile yellow shows second best mAP result. Results outside (parentheses) represent RetinaNet with a ResNet-18 backbone, whereas results inside (parentheses) represent RetinaNet with a ResNet-50 backbone.}
\vspace{-1.5mm}
\label{tab:PASCAL-RAW}
\renewcommand\arraystretch{1.3}
\Large
\resizebox{\linewidth}{!}{%
\begin{tabular}{l|ccccccccc}
\toprule \toprule
\multirow{2}{*}{\textbf{Conditions}}    & \multicolumn{9}{c}{RetinaNet~\cite{lin2017focal_loss} \textit{with} ResNet18 (ResNet50) backbone}                                             \\ \cline{2-10} 
                     & RAW & log-RGB~\cite{log-RGB} & colorimetric~\cite{Colorimetric} & Kar.~\textit{et al.}~\cite{Michael_eccv16} & Lite-ISP~\cite{RAW-to-sRGB_ICCV2021} & InvISP~\cite{invertible_ISP} & Dirty-Pixel~\cite{steven:dirtypixels2021} & Reconfig-ISP~\cite{yu2021reconfigisp} & \textbf{RAW-Adapter} \\ \hline
Normal     &  87.7 (89.2) & \colorbox{red!15}{88.9} (\colorbox{yellow!15}{89.5})  & 87.7  (89.0) & 86.0 (89.4) &  85.2 (89.3) &  84.1 (89.1) & 88.6 (89.0)  &  88.3 (89.4)   &  \colorbox{yellow!15}{88.7} (\colorbox{red!15}{89.6})     \\ \hline
Low-light $\bigstar$  & 80.3 (82.6) & 80.6 (82.3)  & 80.4 (80.9)  &  \colorbox{yellow!15}{81.9} (\colorbox{yellow!15}{84.6})  &    71.9 (73.5)    & 70.9 (74.7)  & 80.8 (83.6)  &  81.5 (83.8)   & \colorbox{red!15}{82.5} (\colorbox{red!15}{86.6})    \\ \hline
Overexposure $\bigstar$ & 87.7 (88.8) & 88.0 (89.1) &  87.4 (88.6)  & 85.6 (86.8)  & 84.2 (85.1) & 86.6 (87.3) & 88.0 (89.0) & \colorbox{yellow!15}{88.2} (\colorbox{yellow!15}{89.1}) &  \colorbox{red!15}{88.7} (\colorbox{red!15}{89.4}) \\ \hline
Flare $\bigstar$ & 84.0 (87.0) & 84.5 (85.2) & 84.2 (86.5)  & 76.4 (77.3) & 80.1 (80.8) & 82.9 (86.9)   &  \colorbox{yellow!15}{85.5} (\colorbox{red!15}{88.9})   & \colorbox{red!15}{85.6} (87.0) & \colorbox{yellow!15}{85.5} (\colorbox{yellow!15}{88.2}) \\ \hline
Low $\&$ Flare $\clubsuit$  & \colorbox{yellow!15}{64.3} (\colorbox{yellow!15}{68.5}) &  64.0 (67.3) &  62.9 (65.6) &  54.8 (58.2) & 58.8 (64.2) & 62.3 (67.9) & \colorbox{red!15}{65.1} (\colorbox{red!15}{69.6}) & 63.2 (65.9) & \colorbox{yellow!15}{64.3} (66.2)  \\ \hline
Fog $\clubsuit$  & 78.5 (79.4) & 78.7 (79.0)  & 78.6 (78.8) &  76.7 (76.8) & 75.6 (74.4) & 78.7 (79.3) &  76.9 (78.0)  & \colorbox{yellow!15}{80.3} (\colorbox{yellow!15}{79.6}) & \colorbox{red!15}{80.5} (\colorbox{red!15}{81.4})   \\ \hline
Rain $\clubsuit$ & 83.6 (86.0)  & 83.9 (85.9) & 83.8 (86.3) &  78.7 (80.1) & 79.0 (80.0)  &  83.9 (\colorbox{red!15}{86.6})  & \colorbox{yellow!15}{84.0} (84.9) & 82.9 (86.3) &  \colorbox{red!15}{84.2} (\colorbox{red!15}{86.9})  \\ \hline
Rain \& Fog $\clubsuit$   & 80.8 (\colorbox{yellow!15}{81.3}) & 80.7 (81.0) & 80.8 (81.0) & 73.5 (74.3) &  79.9 (80.0) & 80.5 (81.2) &  \colorbox{yellow!15}{81.0} (\colorbox{red!15}{81.4}) & 78.5 (80.9) &     \colorbox{red!15}{81.1} (\colorbox{red!15}{81.4})  \\ \hline
Snow $\clubsuit$  & 78.7 (80.5) & \colorbox{yellow!15}{79.5} (80.9) & 79.3 (80.3) & \colorbox{red!15}{80.3} (\colorbox{red!15}{81.3})  &  78.6 (80.5) &  76.4 (80.6)   &  79.0 (\colorbox{yellow!15}{81.1})  & 78.0 (80.7) & \colorbox{yellow!15}{79.5} (80.9)  \\ \hline
Motion Blur $\spadesuit$  & 82.0 (84.7)  & 81.2 (83.9) & \colorbox{yellow!15}{82.1} (84.1) & 81.0 (83.5) &  80.8 (82.7)  & 79.8 (83.3) &  80.7 (85.3) & 82.0 (\colorbox{yellow!15}{85.9}) &  \colorbox{red!15}{82.4} (\colorbox{red!15}{86.4}) \\ \hline
Defocus Blur $\spadesuit$ & \colorbox{yellow!15}{74.5} (79.1)  &  74.2 (78.9) &  73.5 (\colorbox{yellow!15}{79.2})  &  74.0 (76.9) & 72.5 (78.0) & 73.5 (77.5) & 74.3 (78.4) & 73.8 (78.8) &  \colorbox{red!15}{74.6} (\colorbox{red!15}{79.8})  \\ \hline
Sensor Noise $\blacksquare$  & 86.2 (88.0) & 86.2 (87.9) & 86.0 (87.5) & 86.3 (87.9) & 85.8 (87.8) &  85.5 (87.8) & \colorbox{yellow!15}{86.5} (88.4) & \colorbox{red!15}{86.8} (\colorbox{yellow!15}{88.8}) & \colorbox{red!15}{86.8} (\colorbox{red!15}{88.9})    \\ \hline
CMOS-SD $\blacksquare$ & 85.0 (87.1) &  84.8 (87.1) & 85.3 (87.5) & 85.6 (87.5) & 80.3 (84.8) & 84.2 (86.9) & \colorbox{red!15}{86.1} (87.3) & 85.7 (\colorbox{yellow!15}{88.0}) &  \colorbox{yellow!15}{85.9} (\colorbox{red!15}{89.2}) \\ \hline
Vignetting $\blacksquare$ &  85.8 (86.4) &  85.8 (86.2) &   86.4 (86.8) & \colorbox{yellow!15}{87.0} (\colorbox{yellow!15}{88.1}) & 85.1 (85.4)  & 84.3 (86.0) & 86.8 (87.5) & 86.6 (87.8) & \colorbox{red!15}{88.4} (\colorbox{red!15}{89.3}) \\ \hline
Moiré Effect $\blacksquare$ & 85.0 (87.3) &  84.8 (87.0) & 85.2 (86.8) & 83.8 (84.5)  &  83.9 (84.3) &  84.7 (84.8) & 85.5 (\colorbox{yellow!15}{88.7}) & \colorbox{yellow!15}{85.9} (88.3) & \colorbox{red!15}{86.4} (\colorbox{red!15}{88.8}) \\ \hline
CA $\blacksquare$ & \colorbox{yellow!15}{81.9} (84.8) & 81.8 (84.2) & 80.9 (83.7) &  79.5 (84.2) &  77.9 (83.5)  & 76.4 (83.7) &   81.8 (84.9) & \colorbox{red!15}{84.4} (\colorbox{red!15}{85.0}) & \colorbox{red!15}{84.4} (\colorbox{red!15}{85.6}) \\ 
 \bottomrule \bottomrule
\end{tabular}
}
\vspace{-4mm}
\end{table*}

\begin{table*}[]
\caption{Comparison results (mIOU$\%$) on  iPhone XS Max-D dataset. Red background color shows
best mIOU result meanwhile yellow shows second best mIOU result. Results outside (parentheses) represent SegFormer with a MIT-B0 backbone, whereas results inside (parentheses) represent SegFormer with a MIT-B3 backbone.}
\vspace{-1.5mm}
\label{tab:iPhoneXS}
\renewcommand\arraystretch{1.35}
\Large
\resizebox{\linewidth}{!}{%
\begin{tabular}{l|ccccccccc}
\toprule \toprule
\multirow{2}{*}{\textbf{Conditions}}    & \multicolumn{9}{c}{SegFormer~\cite{xie2021segformer} \textit{with} MIT-B0 (MIT-B3) backbone}                                             \\ \cline{2-10} 
                     & RAW & log-RGB~\cite{log-RGB} & colorimetric~\cite{Colorimetric} & Kar.~\textit{et al.}~\cite{Michael_eccv16} & Lite-ISP~\cite{RAW-to-sRGB_ICCV2021} & InvISP~\cite{invertible_ISP} & Dirty-Pixel~\cite{steven:dirtypixels2021} & Reconfig-ISP~\cite{yu2021reconfigisp} & \textbf{RAW-Adapter} \\ \hline
Normal & 57.55 (60.10) & \colorbox{yellow!15}{57.60} (60.24) & 57.54 (60.17) & 56.89 (60.17) & 57.31 (59.94) & 57.52 (60.12)  & \colorbox{yellow!15}{57.60} (60.54) & 57.58 (\colorbox{yellow!15}{60.55}) & \colorbox{red!15}{57.64} (\colorbox{red!15}{60.70}) \\ \hline
Low-light $\bigstar$  & 51.29 (57.34) & 52.23 (57.77) & 51.88 (56.52) &  50.73 (53.42) &  45.85 (50.91) & 40.36 (48.48) & 52.22 (57.45) & \colorbox{yellow!15}{52.33} (\colorbox{yellow!15}{57.78}) & \colorbox{red!15}{52.35} 
(\colorbox{red!15}{58.69}) \\ \hline
Overexposure $\bigstar$ & 58.26 (65.90) & 56.82 (63.43) & 56.73 (63.12) &  57.65 (65.44) & 55.08 (62.82) & 54.69 (62.01) & 57.67 (\colorbox{yellow!15}{66.34}) & \colorbox{yellow!15}{58.29} (65.30) & \colorbox{red!15}{58.58} (\colorbox{red!15}{66.78}) \\ \hline
Flare $\bigstar$ & 55.81 (\colorbox{yellow!15}{63.76}) & 51.93 (63.55) & 52.08 (63.51) & 51.59 (63.01) & 53.26 (63.21) & 51.94 (63.02) & 55.13 (63.36) & \colorbox{yellow!15}{55.99} (63.02) & \colorbox{red!15}{56.33} (\colorbox{red!15}{63.88}) \\ \hline
Low \& Flare $\bigstar$  & 47.14 (\colorbox{yellow!15}{53.42}) & 40.18 (47.71) & 40.23 (47.73) & 38.43 (50.92) & 39.77 (44.31) & 38.84 (46.89) &  47.23 (53.21) &  \colorbox{yellow!15}{47.40} (52.80)  &  \colorbox{red!15}{47.51} (\colorbox{red!15}{53.47}) \\ \hline
Fog $\clubsuit$  & \colorbox{yellow!15}{55.65} (57.29) & 54.49 (58.11) & 53.27 (56.89) &  52.15 (59.24) & 48.94 (56.43) & 53.37 (56.72) & 55.37 (\colorbox{red!15}{62.26}) & 55.48 (60.37) & \colorbox{red!15}{55.66}  (\colorbox{yellow!15}{60.92}) \\ \hline
Rain $\clubsuit$ & 51.88 (\colorbox{yellow!15}{60.44}) & 46.12 (55.21) & 46.03 (53.12) & 44.14 (53.85) & 42.91 (52.94) & 49.61 (57.33) &  51.26 (60.35) & \colorbox{yellow!15}{53.29} (60.32) &  \colorbox{red!15}{53.38} (\colorbox{red!15}{61.86}) \\ \hline
Rain \& Fog $\clubsuit$   & 51.05 (\colorbox{red!15}{58.43}) & 48.77 (56.82) &  48.03 (56.34) & 46.35 (57.75) &  44.72 (55.01) & 51.31 (54.33) &  \colorbox{yellow!15}{51.39} (58.08) & 51.33 (58.09) &  \colorbox{red!15}{51.41} (\colorbox{yellow!15}{58.21})  \\ \hline
Snow $\clubsuit$  & 50.58 (57.83) & 50.65 (56.95)  & 50.54 (56.99) & 47.71 (58.32) & 46.64 (54.83) & 49.06 (55.21) & \colorbox{yellow!15}{50.71} (\colorbox{yellow!15}{58.40}) & 50.66 (58.12) & \colorbox{red!15}{50.94} (\colorbox{red!15}{58.51}) \\ \hline
Motion Blur $\spadesuit$  & 56.06 (63.00) & 55.44 (62.73) & 54.93 (62.62) & 54.92 (61.99) & 49.38 (60.23)  & 48.25 (61.80) & 56.22 (\colorbox{yellow!15}{63.50}) & \colorbox{yellow!15}{56.30} (62.44) & \colorbox{red!15}{56.49} (\colorbox{red!15}{63.60})  \\ \hline
Defocus Blur $\spadesuit$ & 52.31 (59.03) &  52.52 (58.71) &  52.33 (58.97) & 51.57 (55.36) & 45.01 (56.24) & 44.70 (53.31) &  \colorbox{red!15}{52.74} (\colorbox{red!15}{59.55}) & \colorbox{yellow!15}{52.60} (59.00) & 52.35 (\colorbox{yellow!15}{59.09}) \\ \hline
Sensor Noise $\blacksquare$  & 50.46 (58.93) &  \colorbox{yellow!15}{52.16} (\colorbox{yellow!15}{59.20}) &  50.83 (58.87)   & 51.92 (55.92) &  50.88 (56.31) &   49.08 (53.41) &  51.25 (59.19) & 51.87 (59.20) &  \colorbox{red!15}{52.65} (\colorbox{red!15}{59.34}) \\ \hline
CMOS-SD $\blacksquare$ & 56.09 (\colorbox{yellow!15}{64.29})  & 55.81 (63.21) &  54.73 (62.09) & 50.07 (61.33) & 53.21 (63.86) & 50.20 (61.29) & 56.88 (64.16)  & \colorbox{yellow!15}{56.98} (64.08) & \colorbox{red!15}{58.10} (\colorbox{red!15}{64.46}) \\ \hline
Vignetting $\blacksquare$ & \colorbox{red!15}{58.31} (\colorbox{red!15}{66.01}) & 56.87 (64.39) &  56.19 (64.52) & 56.02 (63.87) & 47.22 (63.41) & 49.18 (63.22) & 57.77 (\colorbox{yellow!15}{65.84}) & \colorbox{yellow!15}{58.01} (65.27) &  57.82 (65.55) \\ \hline
Moiré Effect $\blacksquare$ & 53.19 (63.35) &  52.84 (\colorbox{yellow!15}{63.55}) & 52.91 (63.04) & 54.21 (61.87) & 46.87 (61.52) & 44.81 (60.97) & \colorbox{yellow!15}{54.55} (63.24) & 54.38 (63.02) & \colorbox{red!15}{55.01} (\colorbox{red!15}{63.63}) \\ \hline
CA $\blacksquare$ & 53.18 (\colorbox{yellow!15}{62.26}) & 53.51 (62.22) & 53.28 (62.24) & 53.48 (61.81) & 46.81 (59.88) & 47.72 (60.93) &  53.35 (61.93) & \colorbox{yellow!15}{53.68} (62.22) & \colorbox{red!15}{53.76} (\colorbox{red!15}{62.39}) \\  \bottomrule \bottomrule
\end{tabular}
}
\vspace{-4mm}
\end{table*}

\vspace{-2mm}
\subsection{Dataset and Experimental Setting}
\label{sec6-1:setting}

\subsubsection{Dataset}


We conducted object detection and semantic segmentation experiments on \textbf{RAW-Bench} (PASCAL RAW-D, iPhone XS Max-D) and the low-light RAW detection dataset LOD~\cite{LOD_BMVC2021}. An overview of the datasets is provided in Table~\ref{tab:dataset}. For object detection, we used PASCAL RAW-D and LOD~\cite{LOD_BMVC2021}. Here PASCAL RAW~\cite{omid2014pascalraw} is a normal-condition RAW dataset containing 4,259 images captured with a Nikon D3200 DSLR across 3 object classes. Following~\cite{omid2014pascalraw}, we used 2,129 images for training and 2,130 for testing. To assess the robustness of RAW-Adapter and other methods, we synthesized PASCAL RAW-D by introducing various RAW-based corruptions (Sec.\ref{sec4:corruptions}, Fig.\ref{fig:robustness}). LOD~\cite{LOD_BMVC2021} is a real-world low-light dataset with 2,230 RAW images captured using a Canon EOS 5D Mark IV across 8 object classes. Following~\cite{LOD_BMVC2021}, we used 1,800 images for training and 430 for testing.

For semantic segmentation, we used the iPhone XS Max dataset~\cite{RAW_segment_dataset}, a real-world RAW dataset with 1,153 images and corresponding semantic labels. The dataset is split into 806 images for training and 347 for evaluation. Similar to PASCAL RAW-D, we synthesized iPhone XS Max-D by introducing various RAW-based corruptions (Sec.~\ref{sec4:corruptions}).


\subsubsection{Implement Details}

\begin{figure}[t]
    \centering
    \includegraphics[width=1\linewidth]{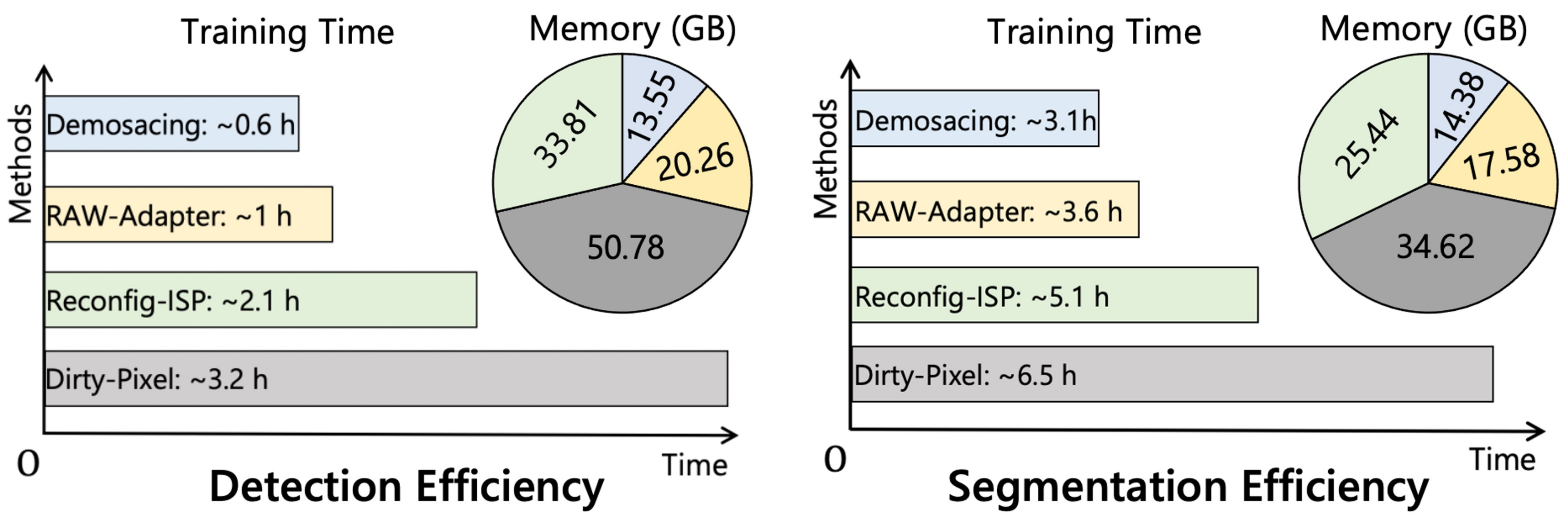}
    \vspace{-6mm}
    \caption{Efficiency comparison with Dirty-Pixel and Reconfig-ISP (Detection: RetinaNet \textit{with} ResNet-18, Segmentation: SegFormer  \textit{with} MIT-B0.).}
    \label{fig:Efficiency}
    \vspace{-3mm}
\end{figure}

We build RAW-Adapter using the open-source computer vision toolboxes \texttt{mmdetection}\cite{chen2019mmdetection} and \texttt{mmsegmentation}\cite{mmseg2020}. Both object detection tasks and semantic segmentation tasks are initialized with ImageNet pre-train weights (see Table.~\ref{tab:dataset}), and we apply the data augmentation pipeline in the default setting, mainly including random crop, random flip, and multi-scale test, etc. For the object detection task, we adopt the 2  mainstream object detectors: RetinaNet~\cite{lin2017focal_loss} and Sparse-RCNN~\cite{sparse_rcnn} with ResNet~\cite{he2016resnet} backbone. For the semantic segmentation task, we choose to use the mainstream segmentation framework Segformer~\cite{xie2021segformer} with their proposed MIT~\cite{xie2021segformer} backbone.

\begin{figure*}[t]
    \centering
    \includegraphics[width=1\linewidth]{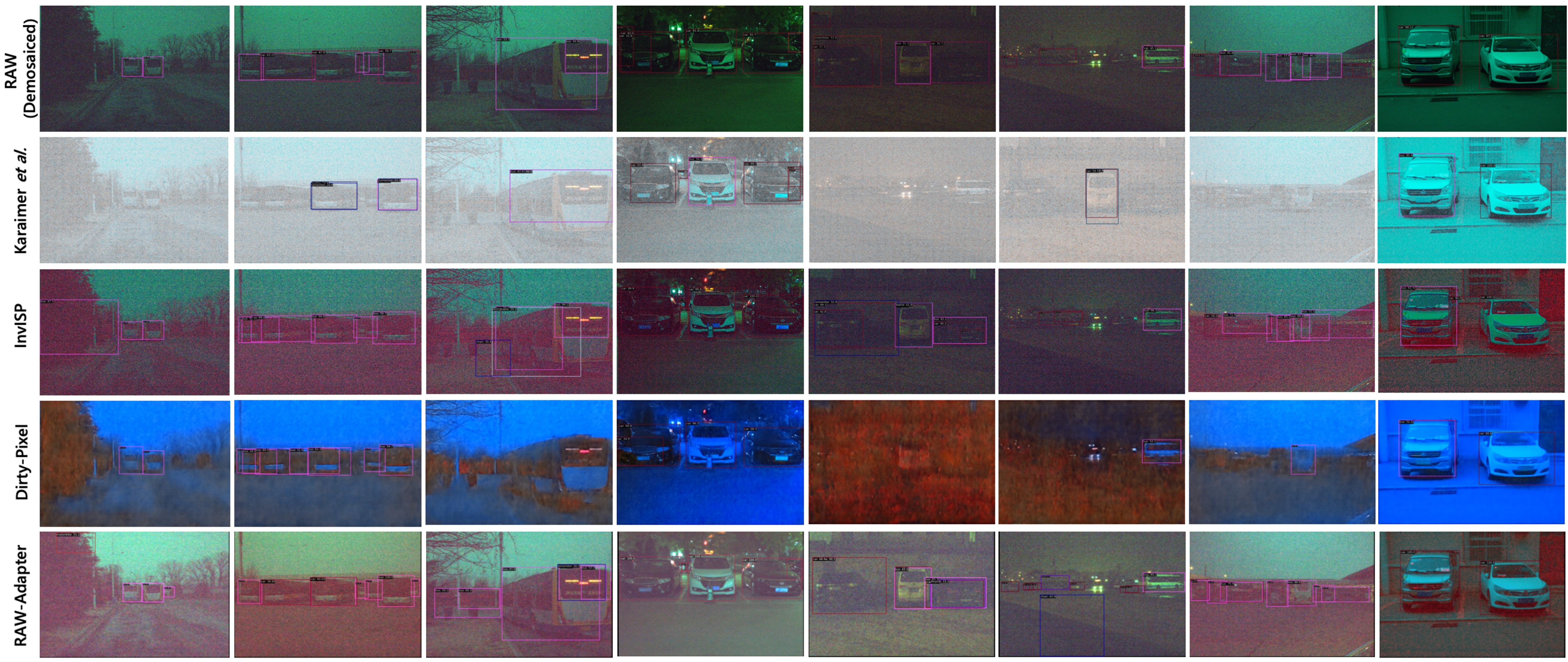}
    \vspace{-7mm}
    \caption{Object detection results on the LOD dataset~\cite{LOD_BMVC2021} using RetinaNet~\cite{lin2017focal_loss} detector with a ResNet-50~\cite{he2016resnet} backbone (please zoom in for details).}
    \label{fig:LOD}
    \vspace{-4mm}
\end{figure*}

\begin{table}
\caption{Comparison results on \textbf{LOD} dataset~\cite{LOD_BMVC2021}. We show the detection performance mAP ($\uparrow$) of RetinaNet (R-Net)~\cite{lin2017focal_loss} and Sparse-RCNN (Sp-RCNN)~\cite{sparse_rcnn}, \textbf{bold} denotes the best
result.}
\label{tab:LOD}
\renewcommand\arraystretch{1.5}
\Huge
\resizebox{\linewidth}{!}{%
\begin{tabular}{c|cccc}
\toprule
\toprule
Methods & RAW  & Default ISP
  & Karaimer~\textit{et al.} 
& InvISP      \\ \hline
mAP(R-Net)      & 58.5 &  58.4  & 54.4   &    56.9                     \\ \hline
 mAP(Sp-RCNN)     & 57.7 & 53.9  &    52.2   &  49.4                    \\ \midrule \midrule
Methods & SID & Dirty-Pixel   &     Reconfig-ISP      & \textbf{RAW-Adapter}     \\ \hline
mAP(R-Net)      & 49.1 & 61.6   & 61.2        & \textbf{62.1}                       \\ \hline
 mAP(Sp-RCNN)      & 43.1  & 58.8         & 58.9         & \textbf{59.2}               \\ \midrule \midrule
 Methods & DNF & \cellcolor[HTML]{ECF4FF}Dirty-Pixel(Aug)   &     \cellcolor[HTML]{ECF4FF}Reconfig-ISP(Aug)      & \cellcolor[HTML]{ECF4FF}\textbf{RAW-Adapter}(Aug)     \\ \hline
mAP(R-Net)      & 48.9 & \cellcolor[HTML]{ECF4FF}61.7($\uparrow$0.1)   & \cellcolor[HTML]{ECF4FF}61.5($\uparrow$0.3)       & \cellcolor[HTML]{ECF4FF}\textbf{62.8}($\uparrow$0.7)                        \\ \hline
 mAP(Sp-RCNN)      & 44.7  & \cellcolor[HTML]{ECF4FF}59.3($\uparrow$0.5)         & \cellcolor[HTML]{ECF4FF}58.9($\uparrow$0.0)         & \cellcolor[HTML]{ECF4FF}\textbf{59.5}($\uparrow$0.3)      \\
 
 \bottomrule
 \bottomrule
\end{tabular}}
\vspace{-3mm}
\end{table}

\subsubsection{Comparison Methods} 

The comparison methods in our work are categorized into three groups: (a) different input image types beyond sRGB, (b) different ISP techniques for converting RAW to sRGB, and (c) state-of-the-art (SOTA) joint-training approaches. A detailed list of these methods is provided in Table.~\ref{tab:comparision}. For group (a), we compare with original demosaiced camera RAW data, log-RGB data generated from RAW~\cite{log-RGB}, and middle colorimetric space representations generated from RAW~\cite{Colorimetric}. For group (b), we compare with different ISP solutions, including the camera default ISP, traditional human handicraft ISP method Karaimer~\textit{et al.}~\cite{Michael_eccv16}, and recent SOTA deep-learning ISP methods Lite-ISP~\cite{RAW-to-sRGB_ICCV2021}, InvISP~\cite{invertible_ISP}, SID~\cite{SID} and DNF~\cite{jincvpr23dnf}, where the SID~\cite{SID} and DNF~\cite{jincvpr23dnf} are specially designed for the low-light condition RAW data. For group (c), we compare with 2 SOTA joint-training methods, Dirty-Pixel~\cite{steven:dirtypixels2021} and Reconfig-ISP~\cite{yu2021reconfigisp}. Meanwhile, to ensure fairness, all competing methods are trained using the same strategy and number of epochs as RAW-Adapter.

\vspace{-1.5mm}
\subsection{Object Detection Evaluation}
\label{sec6-2:detection}


Table~\ref{tab:PASCAL-RAW} presents object detection results on the PASCAL RAW (-D) dataset~\cite{omid2014pascalraw}. We employ the RetinaNet detector~\cite{lin2017focal_loss} with ResNet-18 and ResNet-50 backbones~\cite{he2016resnet}. Our comparisons include different data input types (RAW, log-RGB~\cite{log-RGB}, and colorimetric space~\cite{Colorimetric}), ISP solutions (Karaimer~\textit{et al.}\cite{Michael_eccv16}, Lite-ISP\cite{RAW-to-sRGB_ICCV2021}, and InvISP~\cite{invertible_ISP}), and joint-training methods (Dirty-Pixel~\cite{steven:dirtypixels2021}, Reconfig-ISP~\cite{yu2021reconfigisp}).

As shown in Table~\ref{tab:PASCAL-RAW}, joint-training methods (Dirty-Pixel, Reconfig-ISP, and RAW-Adapter) significantly outperform other approaches, achieving the best performance in most cases. While log-RGB performs well under normal conditions, consistent with~\cite{log-RGB}, deep learning-based ISP methods (Lite-ISP, InvISP) often underperform compared to both RAW images and traditional ISP algorithms~\cite{Michael_eccv16}. Dirty-Pixel~\cite{steven:dirtypixels2021} excels in extreme conditions, such as ``Low-light $\&$ Flare $\bigstar$''.

Overall, RAW-Adapter consistently achieves top-tier performance under both normal and corrupted conditions. Notably, RAW-Adapter with ResNet-18 even outperforms many methods using ResNet-50, particularly in ``Low-light $\bigstar$'' and ``Rain $\clubsuit$'' scenarios. Efficiency comparisons in Fig.\ref{fig:Efficiency} further highlight RAW-Adapter’s advantages in training time and memory. In contrast, Dirty-Pixel is hindered by the heavy computational cost of stacked residual U-Net\cite{UNet} encoders, while Reconfig-ISP suffers from prolonged training due to neural architecture search (NAS) complexity.

Table~\ref{tab:LOD} presents object detection results on the LOD~\cite{LOD_BMVC2021} dataset. We evaluate RetinaNet~\cite{lin2017focal_loss} and Sparse-RCNN~\cite{sparse_rcnn}, both with a ResNet-50 backbone, and compare our results with various ISP methods~\cite{LOD_BMVC2021,Michael_eccv16,invertible_ISP,SID} and joint-training approaches~\cite{steven:dirtypixels2021,yu2021reconfigisp}.
As shown in Table~\ref{tab:LOD}, in low-light scenarios, directly using RAW images outperforms the Default ISP and other ISP methods, including night-specific algorithms like SID~\cite{SID} and DNF~\cite{jincvpr23dnf}. Notably, RAW-Adapter achieves the best overall performance among all tested methods.

Fig.\ref{fig:LOD} shows visualization results, where the background images are RAW data processed by Dirty-Pixel’s pre-encoder\cite{steven:dirtypixels2021} and RAW-Adapter’s input-level adapter. In Table~\ref{tab:LOD}, the blue background indicates results with our RAW-based data augmentation (Sec.~\ref{sec5:augmentation}), which enhances all joint-training methods, with RAW-Adapter benefiting the most.


\vspace{-2mm}
\subsection{Semantic Segmentation Evaluation}
\label{sec6-3:seg}

The semantic segmentation results on iPhone XSmax (-D) dataset are shown in Table.~\ref{tab:iPhoneXS}, here we adopt SegFormer~\cite{xie2021segformer} with different size backbones MIT-B0 and MIT-B3~\cite{xie2021segformer}. Similarly, here we adopt the same comparison methods as in PASCAL RAW-D (Table.~\ref{tab:PASCAL-RAW}). 

 From Table.~\ref{tab:iPhoneXS}, we observe that in the semantic segmentation task, directly using camera RAW as input yields promising results (\textit{i.e.} best in ``Vignetting $\blacksquare$''). Moreover, neither traditional handcrafted ISP method~\cite{Michael_eccv16} nor contemporary deep-learning-based ISP approaches~\cite{invertible_ISP,RAW-to-sRGB_ICCV2021} attain comparable results, possibly because existing ISP algorithms may inadvertently suppress key semantic features under degraded conditions. Overall, the RAW-Adapter consistently outperforms competing methods across standard and degraded conditions, securing 14 of 16 best results with the MIT-B0 backbone and 12 of 16 with the MIT-B3 backbone, thereby underscoring its robust in-domain performance. Fig.~\ref{fig:Efficiency} further demonstrates RAW-Adapter’s computational advantage.
 

\vspace{-2mm}
\subsection{Out-of-domain (OOD) Robustness Evaluation}
\label{sec6-4:OOD}

Beyond the in-domain (ID) experiments described above, we additionally examine the out-of-domain (OOD) robustness of current RAW-based perception methods, including direct RAW input (``Baseline'' in Table.~\ref{tab:PASCALRAW_OOD} and Table.~\ref{tab:iPhoneXS_OOD}), DirtyPixel~\cite{steven:dirtypixels2021}, Reconfig-ISP~\cite{yu2021reconfigisp}, and our RAW-Adapter. Specifically, we train models on normal-condition datasets and evaluate their OOD performance under various common RAW-based corruptions during the inference stage (Fig.~\ref{fig:robustness_info}).

\begin{figure}[t]
    \centering
    \includegraphics[width=1\linewidth]{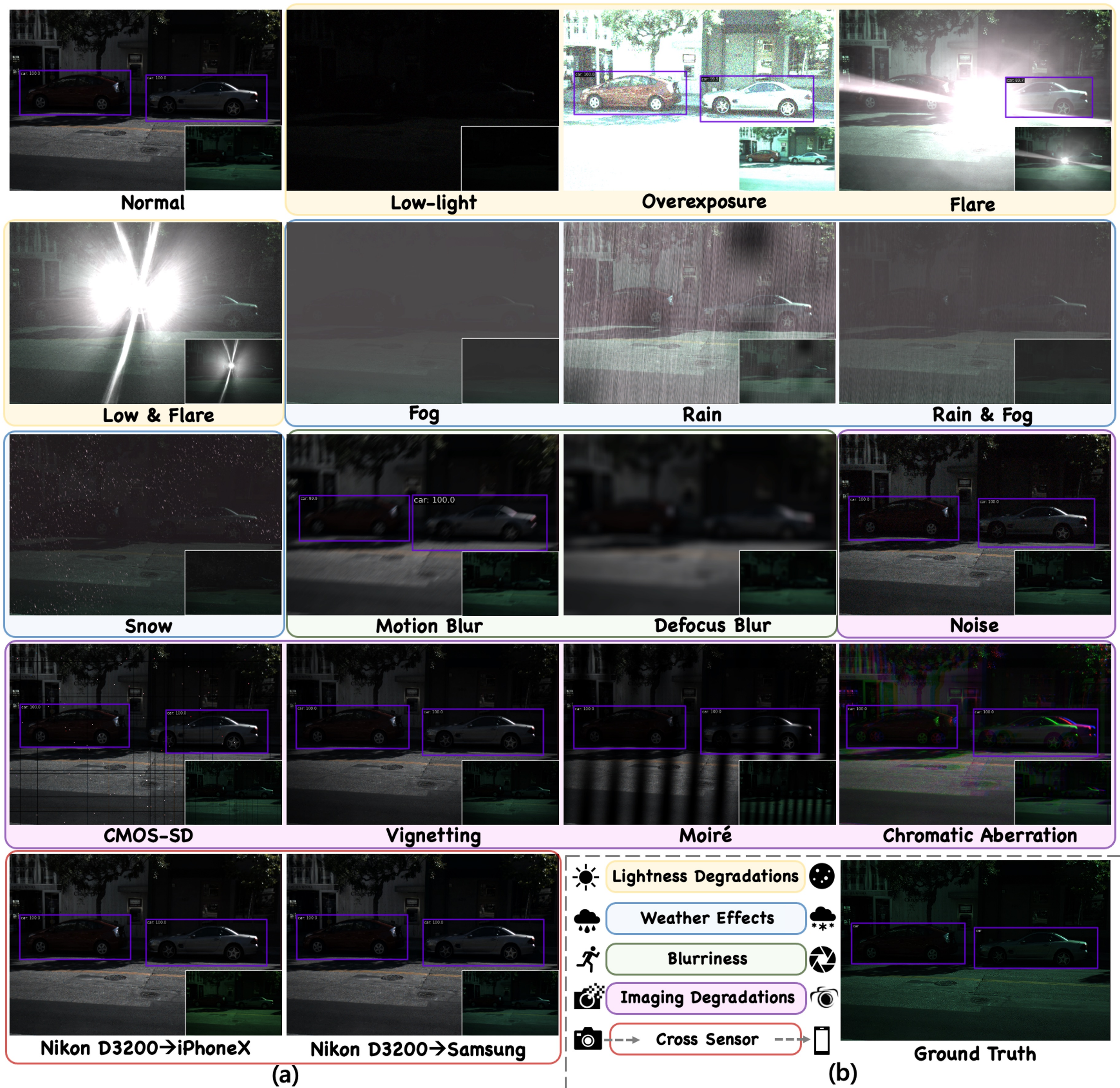}
    \vspace{-6mm}
    \caption{OOD detection results of RAW-Adapter on the PASCAL RAW-D dataset (input RAW at bottom-right of each sub-figure). (b) Ground truth.}
    \label{fig:Det_robust}
    \vspace{-5mm}
\end{figure}

\begin{figure}[t]
    \centering
    \includegraphics[width=1\linewidth]{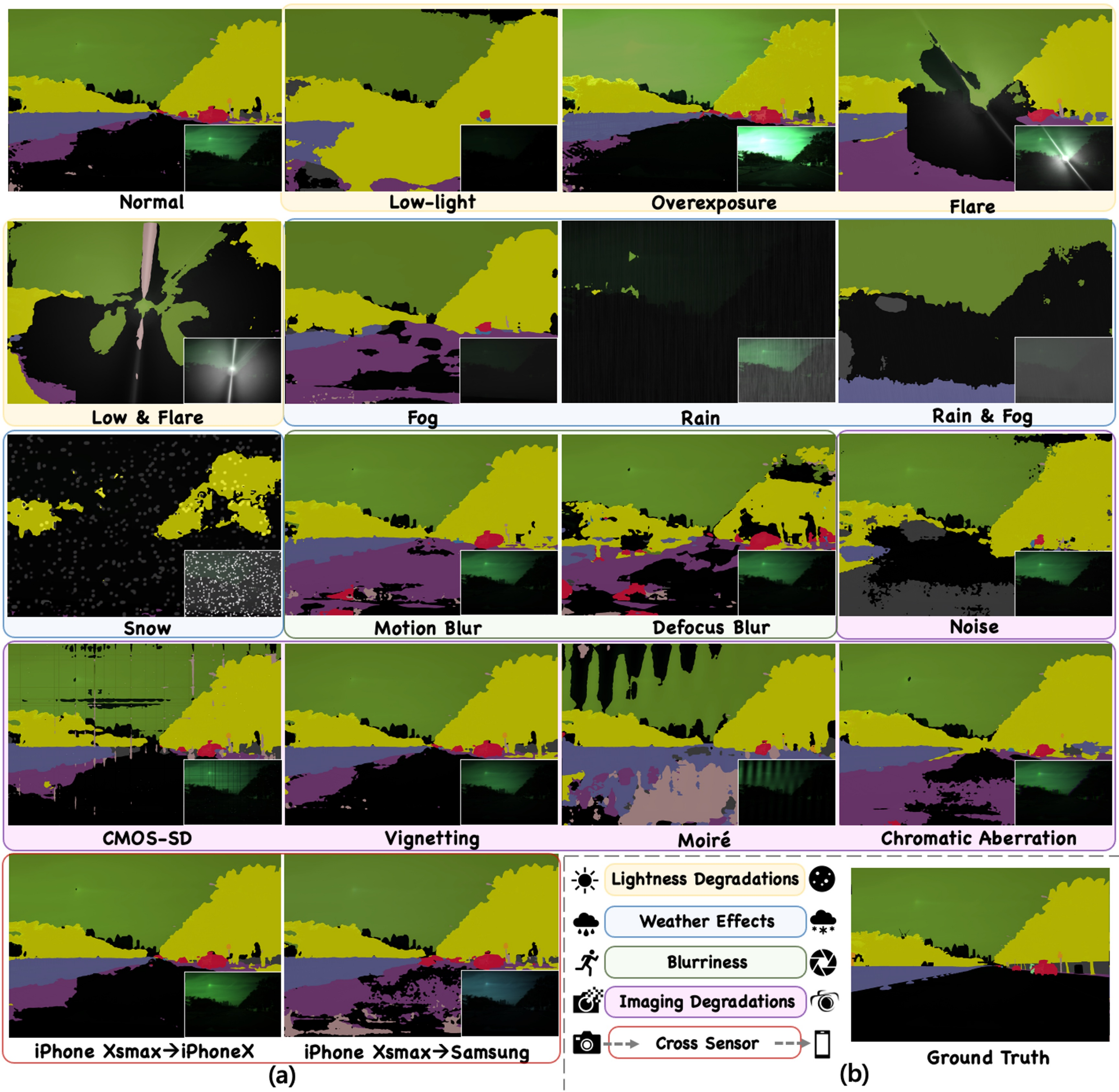}
    \vspace{-6mm}
    \caption{OOD segmentation results of RAW-Adapter on the iPhone XS Max-D dataset (input RAW at bottom-right of each sub-figure). (b) Ground truth.}
    \label{fig:Seg_robust}
    \vspace{-5mm}
\end{figure}

\begin{figure}
    \centering
    
    \includegraphics[width=0.95\linewidth]{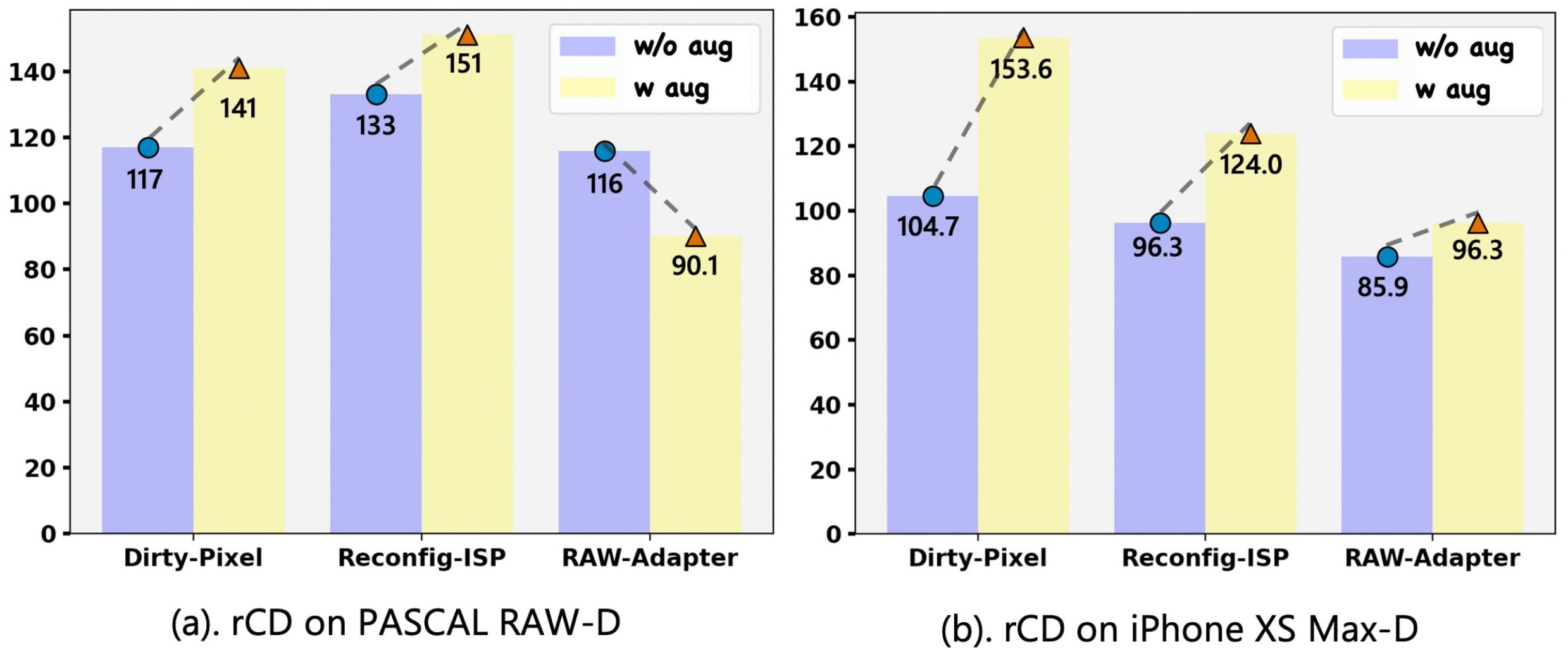}
    \vspace{-1.5mm}
    \caption{The \textit{relative Corruption Degradation} (rCD $\%$) for the \textbf{PASCAL RAW-D detection} and \textbf{iPhone XS Max-D segmentation} tasks. Blue bars represent the original results. Yellow bars represent results achieved with RAW-based data augmentation. The rCD values are computed as the truncated mean across 17 corruption types (excluding the highest and lowest values).}
    \vspace{-2mm}
    \label{fig:rcd}
\end{figure}

\begin{figure}
    
    \centering
    \includegraphics[width=0.95\linewidth]{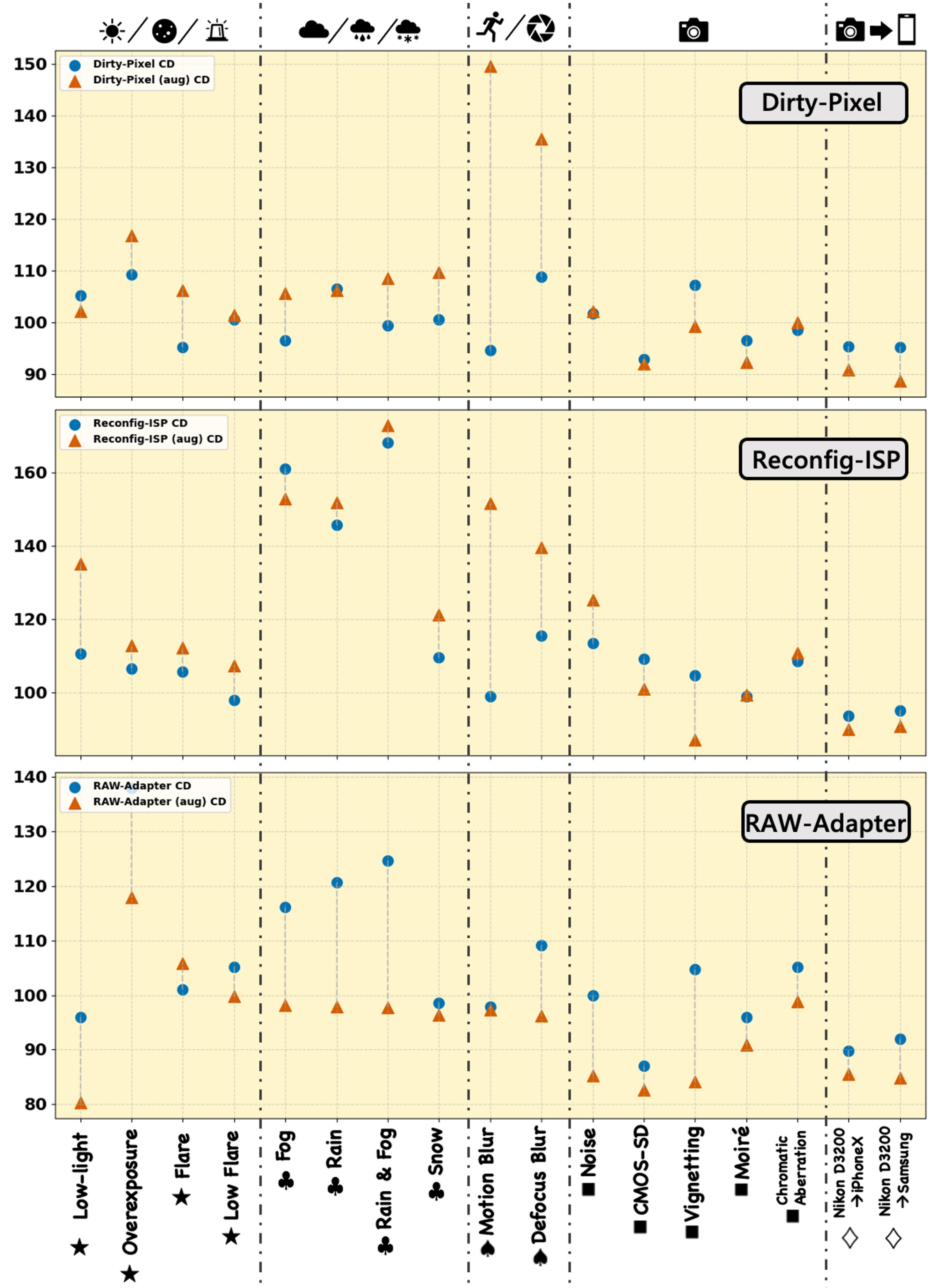}
    \vspace{-3mm}
    \caption{The \textit{Corruption Degradation} (CD $\%$) results on \textbf{PASCAL RAW-D detection} task. Blue circles show original results, yellow triangles show results with RAW-based data augmentation.}
    \vspace{-3mm}
    \label{fig:cd_det}
\end{figure}

\begin{figure}[t]
    \centering
    
    \includegraphics[width=0.95\linewidth]{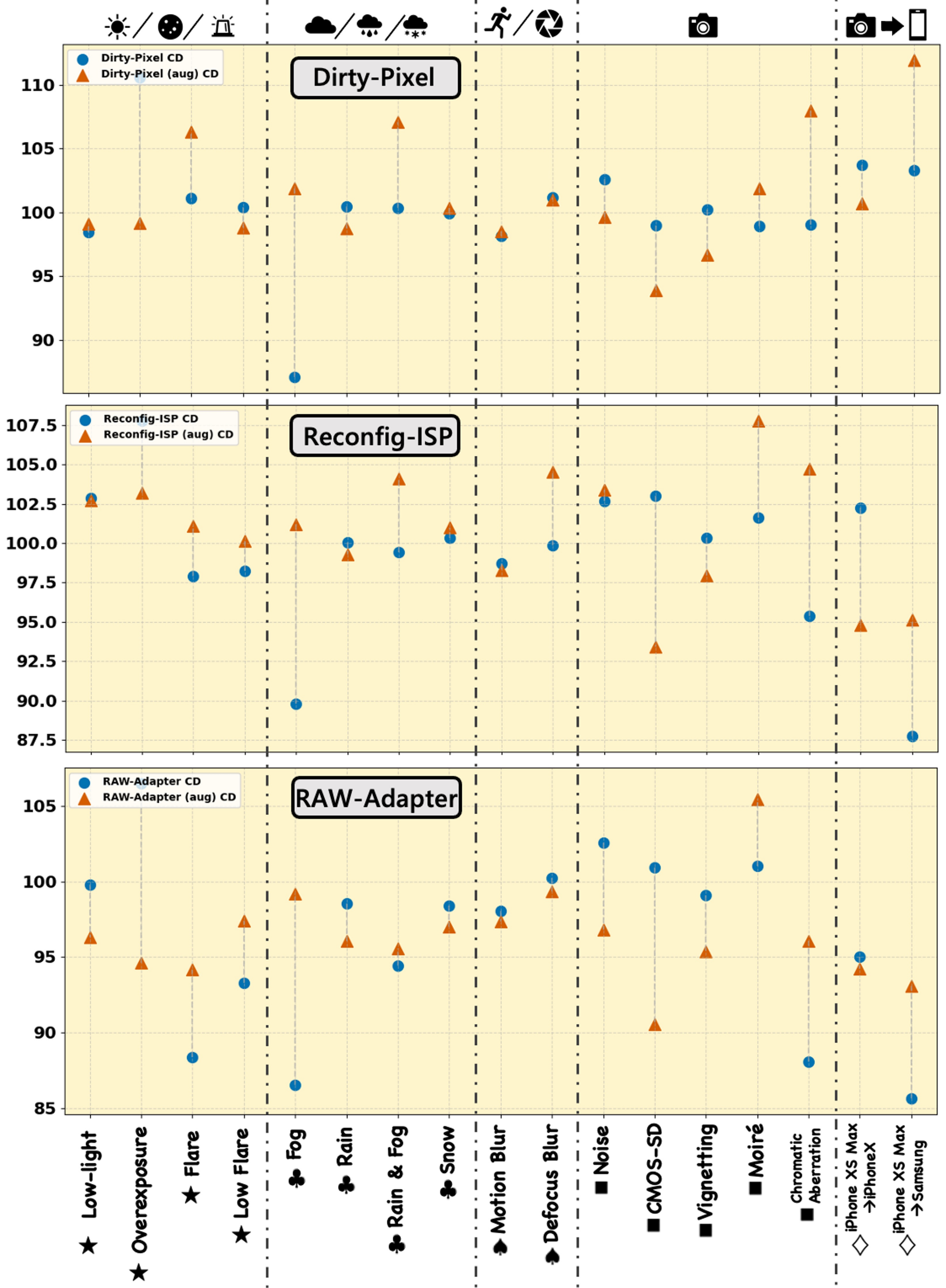}
    \vspace{-3mm}
    \caption{The \textit{Corruption Degradation} (CD $\%$) results on \textbf{iPhone XS Max-D segmentation} task. Blue circles show original results, yellow triangles show results with RAW-based data augmentation.}
    
    \label{fig:cd_seg}
\end{figure}

For OOD robustness evaluation on PASCAL RAW-D, we use  RetinaNet~\cite{lin2017focal_loss} with ResNet-18~\cite{he2016resnet} backbone. Table.~\ref{tab:PASCALRAW_OOD} presents the mAP comparison results. In Table.~\ref{tab:PASCALRAW_OOD}, cells with a white background indicate that the model was trained under normal conditions, while cells with a blue background correspond to models trained using our RAW-based data augmentation (Sec.\ref{sec5:augmentation}) under normal conditions. The corresponding visualization results are shown in Fig.\ref{fig:Det_robust}. For the OOD evaluation on iPhone XS Max-D, we employ SegFormer~\cite{xie2021segformer} with the MIT-B0 backbone. The mIOU comparison results are shown in Table.~\ref{tab:iPhoneXS_OOD}, and the corresponding visualization results are provided in Fig.~\ref{fig:Seg_robust}~\footnote{For full comparison visualizations, please refer to the supplementary part.}. Beyond the mAP and mIOU evaluation metrics, we follow~\cite{Kamann2019BenchmarkingTR_CVPR} and introduce the \textit{Corruption Degradation} (CD) and \textit{relative Corruption Degradation} (rCD) metrics, which are defined as follows:
\begin{equation}
\begin{aligned}
    \rm{CD}_c^{f} & = D_c^{f} / D_c^{ref}\\
    \rm{rCD}_c^{f} & = (D_c^{f} - D_{normal}^{f}) / (D_c^{ref} - D_{normal}^{ref})
\end{aligned}
\label{eq:CD_rCD}
\end{equation}
where $D$ denotes the model error, defined as $D = 1 - mAP$  for object detection tasks and $D = 1 - mIOU$  for semantic segmentation tasks. Additionally, $c$ denotes the corruption type (\textit{i.e.} ``Low-light $\bigstar$''), $f$ denotes the evaluation methods (Dirty-Pixel~\cite{steven:dirtypixels2021}, Reconfig-ISP~\cite{yu2021reconfigisp}, and RAW-Adapter), and $ref$ denotes the reference baseline (direct RAW inputs). The CD results are shown in Fig.~\ref{fig:cd_det} (PASCAL RAW-D) and Fig.~\ref{fig:cd_seg} (iPhone XS Max-D), where we analyze the scores of each joint-training method under different degradation conditions, we predominantly use CD for comparing the robustness of different methods. Meanwhile, the average rCD results under different corruption conditions are shown in Fig.~\ref{fig:rcd}. The CD and rCD values below $100 \%$ indicate better robustness performance compared to the baseline (direct RAW input). The subsequent analysis provides a detailed examination of robustness across different types of corruptions.

\begin{table*}[]
\caption{Out-of-domain (OOD) detection results (mAP $\%$) on  PASCAL RAW-D dataset. We adopt RetinaNet~\cite{lin2017focal_loss} with ResNet-18~\cite{he2016resnet} backbone. ($\bigstar$: lightness degradations, $\clubsuit$: weather effects, $\spadesuit$: blurriness, $\blacksquare$: imaging degradations, $\diamondsuit$: cross sensor). Blue background highlights results obtained by RAW-based data augmentation. \textbf{Bold} denotes best results.}
\label{tab:PASCALRAW_OOD}
\renewcommand\arraystretch{1.15}
\vspace{-1mm}
\Large
\resizebox{\linewidth}{!}{%
\begin{tabular}{c|ccccccccc}
\toprule
\toprule
             & Normal      & Low-light $\bigstar$    & Overexposure $\bigstar$ & Flare $\bigstar$            & Low \& Flare $\bigstar$ & Fog $\clubsuit$   & Rain $\clubsuit$  & Rain \& Fog $\clubsuit$ & Snow $\clubsuit$  \\ \hline
Baseline     &  87.7  &  74.9  & \textbf{80.5}   &  44.4 &   \textbf{17.5}  &  54.7   &   \textbf{58.0}  &  52.8 & 36.3  \\ \hline
\cellcolor[HTML]{ECF4FF}Baseline(Aug)  & \cellcolor[HTML]{ECF4FF}87.1($\downarrow$0.6) & \cellcolor[HTML]{ECF4FF}81.2($\uparrow$6.3) & \cellcolor[HTML]{ECF4FF}\textbf{82.1}($\uparrow$1.6) & \cellcolor[HTML]{ECF4FF}\textbf{50.5}($\uparrow$6.1)  & \cellcolor[HTML]{ECF4FF}23.3($\uparrow$5.8)  & \cellcolor[HTML]{ECF4FF}61.1($\uparrow$6.4) & \cellcolor[HTML]{ECF4FF}66.2($\uparrow$8.2) & \cellcolor[HTML]{ECF4FF}60.9($\uparrow$8.1) & \cellcolor[HTML]{ECF4FF}50.3($\uparrow$14.0) \\ \hline
Dirty-Pixel  & 88.6  & 73.6  &  78.7    &  \textbf{47.1} &   17.1 &  \textbf{56.3} &  55.3   &  \textbf{53.1}  &  36.0 \\ \hline
\cellcolor[HTML]{ECF4FF}Dirty-Pixel(Aug)  & \cellcolor[HTML]{ECF4FF}88.5($\downarrow$0.1) & \cellcolor[HTML]{ECF4FF}80.8($\uparrow$7.2) & \cellcolor[HTML]{ECF4FF}79.1($\uparrow$1.4) & \cellcolor[HTML]{ECF4FF}47.4($\uparrow$0.3)  & \cellcolor[HTML]{ECF4FF}22.2($\uparrow$5.1)  & \cellcolor[HTML]{ECF4FF}58.9($\uparrow$2.6) & \cellcolor[HTML]{ECF4FF}64.1($\uparrow$8.8) & \cellcolor[HTML]{ECF4FF}57.6($\uparrow$4.5) & \cellcolor[HTML]{ECF4FF}45.5($\uparrow$9.5) \\ \hline
Reconfig-ISP &  88.3  &  72.2   &  79.2  & 41.2  &  9.1 &  27.0  &  38.8  &   20.6   &        30.1  \\ \hline
\cellcolor[HTML]{ECF4FF}Reconfig-ISP(Aug)  & \cellcolor[HTML]{ECF4FF}88.3($\uparrow$0.0) & \cellcolor[HTML]{ECF4FF}74.6($\uparrow$2.4) & \cellcolor[HTML]{ECF4FF}79.8($\uparrow$0.6) & \cellcolor[HTML]{ECF4FF}44.4($\uparrow$3.2)  & \cellcolor[HTML]{ECF4FF}17.6($\uparrow$8.5)  & \cellcolor[HTML]{ECF4FF}40.5($\uparrow$13.5) & \cellcolor[HTML]{ECF4FF}48.7($\uparrow$9.9) & \cellcolor[HTML]{ECF4FF}32.4($\uparrow$13.8) & \cellcolor[HTML]{ECF4FF}39.7($\uparrow$9.6) \\ \hline
\textbf{RAW-Adapter}  &  \textbf{88.7} &  \textbf{75.9}  &  63.3  &  43.8  &  13.3     &   47.4  &    49.3  &   41.2 &  \textbf{37.2}     \\
\hline
\cellcolor[HTML]{ECF4FF}\textbf{RAW-Adapter}(Aug)  &  \cellcolor[HTML]{ECF4FF}\textbf{88.9}($\uparrow$0.2) &  \cellcolor[HTML]{ECF4FF}\textbf{84.9}($\uparrow$9.0)  &  \cellcolor[HTML]{ECF4FF}78.9($\uparrow$15.6) &  \cellcolor[HTML]{ECF4FF}47.6($\uparrow$3.8)  &  \cellcolor[HTML]{ECF4FF}\textbf{23.5}($\uparrow$10.2)    & \cellcolor[HTML]{ECF4FF}\textbf{61.8}($\uparrow$14.4) &   \cellcolor[HTML]{ECF4FF}\textbf{66.9}($\uparrow$20.6)  & \cellcolor[HTML]{ECF4FF}\textbf{61.8}($\uparrow$20.6) &  \cellcolor[HTML]{ECF4FF}\textbf{52.1}($\uparrow$14.9)   \\  
\hline \hline
             & Motion Blur $\spadesuit$ & Defocus Blur $\spadesuit$ & Sensor Noise $\blacksquare$  & CMOS-SD $\blacksquare$ & Vignetting $\blacksquare$ & Moiré $\blacksquare$ & CA $\blacksquare$ & iPhoneX $\diamondsuit$ & Samsung $\diamondsuit$ \\ \hline
Baseline     &  57.9 &  \textbf{38.0}  &  \textbf{82.9}    &    80.6  &   \textbf{87.5}  & 80.1 & 73.2     &  87.2       & 87.5 \\ \hline
\cellcolor[HTML]{ECF4FF}Baseline(Aug)  & \cellcolor[HTML]{ECF4FF}77.4($\uparrow$19.5) & \cellcolor[HTML]{ECF4FF}63.4($\uparrow$25.4) & \cellcolor[HTML]{ECF4FF}85.8($\uparrow$2.9) & \cellcolor[HTML]{ECF4FF}82.7($\uparrow$2.1)  & \cellcolor[HTML]{ECF4FF}86.7($\downarrow$0.8)  & \cellcolor[HTML]{ECF4FF}78.2($\downarrow$1.9) & \cellcolor[HTML]{ECF4FF}74.3($\uparrow$1.1) & \cellcolor[HTML]{ECF4FF}86.9($\downarrow$0.6) & \cellcolor[HTML]{ECF4FF}86.8($\downarrow$0.7) \\ \hline
Dirty-Pixel  &  \textbf{60.2}   &   32.5    &    82.6  &   82.0 & 86.6  & 80.8  &  \textbf{73.6}  &  87.8 &   88.1   \\ \hline
\cellcolor[HTML]{ECF4FF}Dirty-Pixel(Aug)  & \cellcolor[HTML]{ECF4FF}66.2($\uparrow$6.0) & \cellcolor[HTML]{ECF4FF}50.4($\uparrow$17.9) & \cellcolor[HTML]{ECF4FF}85.5($\uparrow$2.9) & \cellcolor[HTML]{ECF4FF}84.1($\uparrow$2.1)  & \cellcolor[HTML]{ECF4FF}86.8($\uparrow$0.2)  & \cellcolor[HTML]{ECF4FF}79.9($\downarrow$0.9)& \cellcolor[HTML]{ECF4FF}74.3($\uparrow$0.7) & \cellcolor[HTML]{ECF4FF}88.1($\uparrow$0.3) & \cellcolor[HTML]{ECF4FF}88.3($\uparrow$0.2) \\ \hline
Reconfig-ISP &  58.3  &  28.4 &  80.6  &  78.8   &   86.9  &   80.3  &  70.9      &  88.0  &  88.1 \\ \hline
\cellcolor[HTML]{ECF4FF}Reconfig-ISP(Aug)  & \cellcolor[HTML]{ECF4FF}65.7($\uparrow$7.4) & \cellcolor[HTML]{ECF4FF}48.9($\uparrow$20.5) & \cellcolor[HTML]{ECF4FF}82.2($\uparrow$1.6) & \cellcolor[HTML]{ECF4FF}82.5($\uparrow$3.7)  & \cellcolor[HTML]{ECF4FF}88.4($\uparrow$1.5)  & \cellcolor[HTML]{ECF4FF}78.3($\downarrow$2.0) & \cellcolor[HTML]{ECF4FF}71.5($\uparrow$0.6) & \cellcolor[HTML]{ECF4FF}88.2($\uparrow$0.2) & \cellcolor[HTML]{ECF4FF}88.0($\downarrow$0.1) \\ \hline
\textbf{RAW-Adapter}  &  58.8  &  32.3   &  \textbf{82.9} & \textbf{83.1}  & 86.9  & \textbf{80.9} &  71.8  & \textbf{88.5}  & \textbf{88.5}  \\ \hline
\cellcolor[HTML]{ECF4FF}\textbf{RAW-Adapter}(Aug)  & \cellcolor[HTML]{ECF4FF}\textbf{78.0}($\uparrow$19.2)  & \cellcolor[HTML]{ECF4FF}\textbf{64.8}($\uparrow$32.5)  & \cellcolor[HTML]{ECF4FF}\textbf{87.8}($\uparrow$4.9) & \cellcolor[HTML]{ECF4FF}\textbf{85.7}($\uparrow$2.6) & \cellcolor[HTML]{ECF4FF}\textbf{88.8}($\uparrow$1.9) & \cellcolor[HTML]{ECF4FF}\textbf{80.2}($\downarrow$0.7) & \cellcolor[HTML]{ECF4FF}\textbf{74.6}($\uparrow$2.8) & \cellcolor[HTML]{ECF4FF}\textbf{88.8}($\uparrow$0.3) & \cellcolor[HTML]{ECF4FF}\textbf{88.8}($\uparrow$0.3) \\  \bottomrule \bottomrule
\end{tabular}
}
\vspace{-4mm}
\end{table*}


\textbf{Performance \textit{w.r.t} normal condition:} Table~\ref{tab:PASCALRAW_OOD} shows that RAW-based data augmentation increases RAW-Adapter's performance on PASCAL RAW detection by 0.2 points. It does not benefit the other two joint-training methods and lowers the baseline's performance. Table~\ref{tab:iPhoneXS_OOD} indicates that augmentation improves iPhone XS Max-D segmentation results for all three joint-training methods, while the baseline declines, likely due to its lack of adaptive parameters for augmentation.


\begin{table*}[h]
\caption{Out-of-domain (OOD) segmentation results (mIOU $\%$) on iPhone XS Max-D dataset. We adopt SegFormer~\cite{xie2021segformer} with MIT-B0~\cite{xie2021segformer} backbone. ($\bigstar$: lightness degradations, $\clubsuit$: weather effects, $\spadesuit$: blurriness, $\blacksquare$: imaging degradations, $\diamondsuit$: cross sensor). Blue background highlights results obtained by RAW-based data augmentation. \textbf{Bold} denotes best results.}
\label{tab:iPhoneXS_OOD}
\renewcommand\arraystretch{1.15}
\vspace{-1mm}
\Large
\resizebox{\linewidth}{!}{%
\begin{tabular}{c|ccccccccc}
\toprule
\toprule
             & Normal      & Low-light $\bigstar$    & Overexposure $\bigstar$ & Flare $\bigstar$            & Low \& Flare $\bigstar$ & Fog $\clubsuit$   & Rain $\clubsuit$  & Rain \& Fog $\clubsuit$ & Snow $\clubsuit$  \\ \hline
Baseline     &  57.55  & 29.65 & \textbf{54.93} & 22.02  & 6.12 & 29.65 &  4.12  & 7.82  &  15.53 \\ \hline
\cellcolor[HTML]{ECF4FF}Baseline(Aug)  & \cellcolor[HTML]{ECF4FF}56.46($\downarrow$1.09) & \cellcolor[HTML]{ECF4FF}48.08($\uparrow$18.43) & \cellcolor[HTML]{ECF4FF}55.94($\uparrow$1.01) &  \cellcolor[HTML]{ECF4FF}31.10($\uparrow$9.08) & \cellcolor[HTML]{ECF4FF}9.93($\uparrow$3.81)  & \cellcolor[HTML]{ECF4FF}42.08($\uparrow$12.43) & \cellcolor[HTML]{ECF4FF}17.76($\uparrow$13.64) & \cellcolor[HTML]{ECF4FF}21.16($\uparrow$13.34) & \cellcolor[HTML]{ECF4FF}22.59($\uparrow$7.06)  \\ \hline
Dirty-Pixel  & 57.60 & \textbf{30.73} & 50.17 & 21.13 & 5.72 &  38.74 & 3.69 & 7.50 &  15.58 \\ \hline

\cellcolor[HTML]{ECF4FF}Dirty-Pixel(Aug)  & \cellcolor[HTML]{ECF4FF}58.04($\uparrow$1.44) & \cellcolor[HTML]{ECF4FF}48.53($\uparrow$17.80) & \cellcolor[HTML]{ECF4FF}56.32($\uparrow$6.15) &  \cellcolor[HTML]{ECF4FF}26.76($\uparrow$5.63) & \cellcolor[HTML]{ECF4FF}11.01($\uparrow$5.29)  & \cellcolor[HTML]{ECF4FF}40.98($\uparrow$2.24) & \cellcolor[HTML]{ECF4FF}18.79($\uparrow$15.10) & \cellcolor[HTML]{ECF4FF}15.56($\uparrow$8.06) & \cellcolor[HTML]{ECF4FF}22.33($\uparrow$6.75)  \\ \hline
Reconfig-ISP & 57.58 & 27.65  & 51.43 & 23.65 & 7.77 & 36.82  & 4.06 & 8.35 &  15.26  \\ \hline
\cellcolor[HTML]{ECF4FF}Reconfig-ISP(Aug)  & \cellcolor[HTML]{ECF4FF}57.99($\uparrow$0.41) & \cellcolor[HTML]{ECF4FF}46.68($\uparrow$19.03) & \cellcolor[HTML]{ECF4FF}54.53($\uparrow$3.10) & \cellcolor[HTML]{ECF4FF}30.33($\uparrow$6.68)  &  \cellcolor[HTML]{ECF4FF}9.82($\uparrow$2.05) & \cellcolor[HTML]{ECF4FF}41.38($\uparrow$4.56) & \cellcolor[HTML]{ECF4FF}18.35($\uparrow$14.29) & \cellcolor[HTML]{ECF4FF}17.92($\uparrow$9.57) & \cellcolor[HTML]{ECF4FF}21.83($\uparrow$6.57) \\ \hline
\textbf{RAW-Adapter}  & \textbf{57.64}  & 29.79 & 52.01 & \textbf{31.09} & \textbf{12.45} & \textbf{39.13} & \textbf{5.53} & \textbf{12.98}  & \textbf{16.88} 
 \\ \hline
\cellcolor[HTML]{ECF4FF}\textbf{RAW-Adapter}(Aug) & \cellcolor[HTML]{ECF4FF}\textbf{58.58}($\uparrow$0.94) & \cellcolor[HTML]{ECF4FF}\textbf{49.99}($\uparrow$20.20) & \cellcolor[HTML]{ECF4FF}\textbf{58.32}($\uparrow$6.31)  & \cellcolor[HTML]{ECF4FF}\textbf{35.12}($\uparrow$4.03) & \cellcolor[HTML]{ECF4FF}\textbf{12.28}($\downarrow$0.17) & \cellcolor[HTML]{ECF4FF}\textbf{42.56}($\uparrow$3.43)  & \cellcolor[HTML]{ECF4FF}\textbf{21.02}($\uparrow$15.49) & \cellcolor[HTML]{ECF4FF}\textbf{24.65}($\uparrow$11.67) & \cellcolor[HTML]{ECF4FF}\textbf{24.91}($\uparrow$8.03)
\\ \hline \hline
             & Motion Blur $\spadesuit$ & Defocus Blur $\spadesuit$ & Sensor Noise $\blacksquare$  & CMOS-SD $\blacksquare$ & Vignetting $\blacksquare$ & Moiré $\blacksquare$ & CA $\blacksquare$ & iPhoneX $\diamondsuit$ & Samsung $\diamondsuit$ 
             \\ \hline
Baseline     & 52.16  &  37.76 &  \textbf{32.42}  & 34.42 & 56.23 & 37.79 & 34.40   &  54.21 & 48.80 \\ \hline
\cellcolor[HTML]{ECF4FF}Baseline(Aug)  & \cellcolor[HTML]{ECF4FF}54.97($\uparrow$2.81) & \cellcolor[HTML]{ECF4FF}49.71($\uparrow$11.95) & \cellcolor[HTML]{ECF4FF}50.60($\uparrow$18.18) &  \cellcolor[HTML]{ECF4FF}36.90($\uparrow$2.48) & \cellcolor[HTML]{ECF4FF}55.88($\downarrow$0.35)  & \cellcolor[HTML]{ECF4FF}\textbf{34.77}($\downarrow$3.02) & \cellcolor[HTML]{ECF4FF}46.38($\uparrow$11.98) & \cellcolor[HTML]{ECF4FF}55.79($\uparrow$1.58) & \cellcolor[HTML]{ECF4FF}54.74($\uparrow$5.94)  \\ \hline
Dirty-Pixel  & 53.05 & 37.04 & 30.66 &  35.09 & 56.14 & \textbf{38.45}  & 35.02 & 52.52 & 47.12  \\ \hline
\cellcolor[HTML]{ECF4FF}Dirty-Pixel(Aug)  & \cellcolor[HTML]{ECF4FF}55.65($\uparrow$2.60) & \cellcolor[HTML]{ECF4FF}49.21($\uparrow$12.17) & \cellcolor[HTML]{ECF4FF}50.78($\uparrow$20.12) &  \cellcolor[HTML]{ECF4FF}40.75($\uparrow$5.66) & \cellcolor[HTML]{ECF4FF}57.34($\uparrow$1.20)  & \cellcolor[HTML]{ECF4FF}33.55($\downarrow$4.90) & \cellcolor[HTML]{ECF4FF}42.10($\uparrow$7.08) & \cellcolor[HTML]{ECF4FF}55.47($\uparrow$2.95) & \cellcolor[HTML]{ECF4FF}49.33($\uparrow$2.21)  \\ \hline
Reconfig-ISP & 52.78 & \textbf{37.84} & 30.61 & 32.44   & 56.08 & 36.79 & 37.42 & 53.19 & 55.07 \\ \hline
\cellcolor[HTML]{ECF4FF}Reconfig-ISP(Aug)  & \cellcolor[HTML]{ECF4FF}55.74($\uparrow$2.96) & \cellcolor[HTML]{ECF4FF}47.42($\uparrow$9.58) & \cellcolor[HTML]{ECF4FF}48.92($\uparrow$18.31) &  \cellcolor[HTML]{ECF4FF}41.05($\uparrow$8.61) & \cellcolor[HTML]{ECF4FF}56.79($\uparrow$0.71)  & \cellcolor[HTML]{ECF4FF}29.71($\downarrow$7.08) & \cellcolor[HTML]{ECF4FF}43.84($\uparrow$6.42) & \cellcolor[HTML]{ECF4FF}58.09($\uparrow$4.90) & \cellcolor[HTML]{ECF4FF}56.94($\uparrow$1.87)
 \\ \hline
\textbf{RAW-Adapter}  & \textbf{53.09} & 37.62 & 30.70 &  33.81 & \textbf{56.64}  & 37.17 & \textbf{42.24} & \textbf{56.50} & \textbf{56.16} 

 \\ \hline
\cellcolor[HTML]{ECF4FF}\textbf{RAW-Adapter}(Aug) & \cellcolor[HTML]{ECF4FF}\textbf{56.16}($\uparrow$3.07) & \cellcolor[HTML]{ECF4FF}\textbf{50.05}($\uparrow$12.43) & \cellcolor[HTML]{ECF4FF}\textbf{52.19}($\uparrow$21.49) & \cellcolor[HTML]{ECF4FF}\textbf{42.87}($\uparrow$9.06) & \cellcolor[HTML]{ECF4FF}\textbf{57.92}($\uparrow$1.28) & \cellcolor[HTML]{ECF4FF}31.22($\downarrow$5.95) &  \cellcolor[HTML]{ECF4FF}\textbf{48.50}($\uparrow$6.26) & \cellcolor[HTML]{ECF4FF}\textbf{58.34}($\uparrow$1.84) & \cellcolor[HTML]{ECF4FF}\textbf{57.87}($\uparrow$1.71)
\\ \bottomrule \bottomrule
\end{tabular}
}
\vspace{-4mm}
\end{table*}

\textbf{Performance \textit{w.r.t} lightness degradations:}
In Tables~\ref{tab:PASCALRAW_OOD} and \ref{tab:iPhoneXS_OOD}, the $\bigstar$ symbol denotes performance under various lightness degradations. In Table~\ref{tab:PASCALRAW_OOD} (white background), “Low-light $\&$ Flare $\bigstar$” has the greatest negative impact on detection performance. The combined effects of flare and low-light create challenging conditions (see Fig.\ref{fig:Det_robust}). Baseline methods remain robust against lighting variations; RAW-Adapter struggles with “Overexposure $\bigstar$” yet performs well under “Low-light $\bigstar$”. With data augmentation (blue background), all models improve; Fig.\ref{fig:cd_det} shows that RAW-Adapter’s CD value decreases (indicating better robustness), while CD values for Dirty-Pixel and Reconfig-ISP increase.

In Table~\ref{tab:iPhoneXS_OOD} (white background) for semantic segmentation, “Low-light $\&$ Flare $\bigstar$” also has the most severe impact, whereas “Overexposure $\bigstar$” has the least (see Fig.~\ref{fig:Seg_robust}). Data augmentation (blue background) enhances performance for all models, with RAW-Adapter (Aug) achieving all best results.


\textbf{Performance \textit{w.r.t} weather effects:} The $\clubsuit$ symbol denotes the model performance under various weather corruptions. In Table~\ref{tab:PASCALRAW_OOD} (white background), ``Snow $\clubsuit$'' and ``Rain $\&$ Fog $\clubsuit$'' lead to the largest drop in detection performance, likely due to object occlusion causing mispredictions (see Fig.~\ref{fig:Det_robust}). After data augmentation (blue background), all methods improved, with RAW-Adapter achieving the best results.

Meanwhile, in Table~\ref{tab:iPhoneXS_OOD} (white background) for semantic segmentation, ``Rain $\clubsuit$'' and ``Rain $\&$ Fog $\clubsuit$'' cause the most significant decline, largely because rain streaks disrupt the semantic information in RAW data (see Fig.~\ref{fig:Seg_robust}). After data augmentation, performance improved across all methods. Notably, RAW-Adapter consistently delivered the best performance regardless of augmentation, with its CD $\%$ remaining below 100 $\%$ (see Fig.~\ref{fig:cd_seg}).





\textbf{Performance \textit{w.r.t} blurriness:} The $\spadesuit$ symbol denotes performance under blurry corruptions. In Table~\ref{tab:PASCALRAW_OOD} (white background), ``Defocus Blur $\spadesuit$'' severely impacts all methods (e.g., RAW-Adapter drops from 88.7 to 32.3). After applying data augmentation (blue background), all methods improved, partly due to the RAW-based data augmentation simulating diverse degradation conditions and increasing the number of degradation samples during training.


In Table~\ref{tab:iPhoneXS_OOD} (white background) for semantic segmentation, consistent with the detection results, ``Defocus Blur $\spadesuit$'' continues to have a more severe impact than ``Motion Blur $\spadesuit$''. This partly because defocus blur leads to a greater loss of fine texture details~\cite{Kamann2019BenchmarkingTR_CVPR}, which are crucial for accurate segmentation (see Fig.~\ref{fig:Seg_robust}). 

\textbf{Performance \textit{w.r.t} camera imaging degradations:} The $\blacksquare$ symbol denotes the model performance under various camera imaging degradations. In Table~\ref{tab:PASCALRAW_OOD} (white background), ``Chromatic Aberration (CA) $\blacksquare$'' causes the significant performance drop for detection. Meanwhile, RAW-Adapter maintains strong out-of-domain robustness under ``Noise $\blacksquare$'', ``CMOS-SD $\blacksquare$'', and ``Moiré Effect $\blacksquare$''. Interestingly, after data augmentation (blue background), all models show improved performance under camera imaging degradations, except for ``Moiré Effect $\blacksquare$'', which performs even worse after RAW-based data augmentation. Additionally, RAW-Adapter's CD$\%$ values remaining below 100$\%$ indicate that it exhibits better robustness than the other two methods (see Fig.~\ref{fig:cd_det}).

Meanwhile, in Table~\ref{tab:iPhoneXS_OOD} (white background) for semantic segmentation, ``Noise $\blacksquare$'' and ``CMOS-SD $\blacksquare$'' cause significant performance drops. Fig.~\ref{fig:Seg_robust} shows that these two types of degradations severely affect semantic information. Similar to the conclusion in detection, after data augmentation (blue background), only the ``Moiré Effect $\blacksquare$'' exhibits a performance drop.


\textbf{Performance \textit{w.r.t} cross sensor ability:} The $\diamondsuit$ symbol denotes the model performance under different camera color response functions (cross-sensor scenario). Specifically, it represents the case where the model is trained on RAW data from camera sensor A and tested on RAW data from camera sensor B. In the object detection task (Table~\ref{tab:PASCALRAW_OOD}), the PASCAL RAW dataset~\cite{omid2014pascalraw} was collected using a Nikon D3200 DSLR camera, and a slight performance drop is observed under the ``Nikon D3200 $\rightarrow$ iPhoneX $\diamondsuit$'' and ``Nikon D3200 $\rightarrow$ Samsung S9 $\diamondsuit$'' settings. Interestingly, after RAW-based data augmentation (blue background in Table~\ref{tab:PASCALRAW_OOD}), only the baseline method's performance decreases overall, while the performance of the other joint-training methods generally improves.

For the semantic segmentation task (Table.~\ref{tab:iPhoneXS_OOD}), the iPhone XS Max dataset~\cite{RAW_segment_dataset} was collected using the iPhone XS Max phone camera. Because the iPhone XS Max and iPhone X have very similar camera models and sensor types~\footnote{https://www.apple.com/au/iphone/compare/?modelList=iphone-xs,iphone-x,iphone-xs-max}, there is little OOD performance drop observed in ``iPhone XS Max $\rightarrow$ iPhoneX $\diamondsuit$'' setting (e.g., Baseline from 57.55 to 54.21). However, in the “iPhone XS Max $\rightarrow$ Samsung S9 $\diamondsuit$” setting, the performance drop is much more significant (e.g., Baseline from 57.55 to 48.80), due to the substantial differences in sensor characteristics and ISP. Overall, the RAW-based data augmentation (blue background in Table~\ref{tab:iPhoneXS_OOD}) can effectively enhance the segmentation performance of all methods.

\textbf{Overall Discussion:} From Fig.\ref{fig:cd_det} and Fig.\ref{fig:cd_seg}, we can observe that RAW-Adapter's CD values are generally lower, indicating better robustness compared to other methods. Notably, after data augmentation, although most methods show performance improvements, the robustness of Dirty-Pixel and Reconfig-ISP tends to degrade relative to the baseline (CD values larger than 100 $\%$). Meanwhile, RAW-Adapter demonstrates a robustness improving trend (CD value decrease). The rCD comparison results in Fig.~\ref{fig:rcd} also indicates RAW-Adapter's superior overall robustness.

RAW-Adapter performs less optimally in certain cases, such as ``Overexposure $\bigstar$'' and ``Flare $\bigstar$'' in object detection and ``Moire Effects'' in semantic segmentation. However, overall, RAW-Adapter achieves strong results across most scenarios. Notably, after data augmentation, RAW-Adapter achieves the best performance in 12 out of 18 OOD detection cases and 17 out of 18 OOD segmentation cases.

    

\section{Conclusion}

In this work, we introduce \textbf{RAW-Adapter}, which bridges pre-trained sRGB models and RAW images using both input- and model-level adapters. Our approach achieves state-of-the-art (SOTA) performance over existing ISP and joint-training methods and further benefits from RAW-based data augmentation. Additionally, we propose \textbf{RAW-Bench}, which establish a robust benchmark covering lighting, weather, blur, and imaging degradations and cross-sensor performance.

For future research directions, we believe that exploring multi-task architectures based on camera RAW images is a highly meaningful direction. An effective multi-task system can facilitate the seamless integration of various visual tasks on RAW images within large-scale models. Additionally, incorporating RAW-Adapter into existing end-to-end autonomous driving systems to fully leverage its advantages in real-world scenarios is another promising and impactful direction. By harnessing the strengths of RAW images to counteract various lightness and weather degradations encountered in self-driving cars, this approach may lead transformative advancements in future autonomous driving systems.

\bibliographystyle{IEEEtran}
\bibliography{references}

\clearpage

\title{Supplementary Material for RAW-Adapter: Adapting Pre-trained Visual Model to Camera RAW Images and A Benchmark}

\author{Ziteng Cui, Jianfei Yang, Tatsuya Harada
\thanks{Ziteng Cui is with the RCAST, The University of Tokyo, Japan.}
\thanks{Jianfei Yang is with the MARS Lab at Nanyang Technological University, Singapore}
\thanks{Tatsuya Harada is with the RCAST, The University of Tokyo and RIKEN AIP, Japan.}
\thanks{\Envelope \ Corresponding Author: Tatsuya Harada (harada@mi.t.u-tokyo.ac.jp)}}

\maketitle

\renewcommand\thesection{\Alph{section}}
\renewcommand\thesubsection{\thesection.\arabic{subsection}}
\renewcommand\thetable{\Alph{section}\arabic{table}} 
\renewcommand\thefigure{\Alph{section}\arabic{figure}}

The supplementary material includes the following sections: In Sec.~\ref{sec1_ISP_revisited}, we revisit the image signal processor (ISP) pipeline. In Sec.~\ref{sec2_Ablation}, we present additional ablation studies, including the impact of different blocks in RAW-Adapter and the effects of various components in RAW-based data augmentation. In Sec.~\ref{sec3_Detail_Design}, we detail the design and hyperparameter settings of the input-level adapter. In Sec.~\ref{sec4_ade20k_raw}, we provide semantic segmentation results on our synthesized ADE20K-RAW dataset. Finally, in Sec.~\ref{sec5_visulization}, we present additional visualization results on the PASCAL RAW-D dataset and the iPhone XS Max-D dataset (Table.~\uppercase\expandafter{\romannumeral6} and Table.~\uppercase\expandafter{\romannumeral7} in main text).

\section{ISP revisited}
\label{sec1_ISP_revisited}
Here we give a brief review of the digital camera image formation process, from camera sensor RAW data to output rendered sRGB. Please refer to Fig.2 (a) in main paper for an illustrative diagram, the basic ISP steps mainly include:

\textit{(a). Pre-processing} involves some pre-process operations such as BlackLevel adjustment, WhiteLevel adjustment, and lens shading correction.

\textit{(b). Noise reduction} eliminates noise and keeps the visual quality of image, this step is closely related to exposure time and camera ISO settings~\cite{Wei_2020_CVPR,Dancing_under_light}.

\textit{(c). Demosaicing} is used to reconstruct a 3-channel color image from a single-channel RAW, executed through interpolation of the absent values in the Bayer pattern, relying on neighboring values in the CFA.

\textit{(d). White Balance}  simulates the color constancy of human visual system (HVS). An auto white balance (AWB) algorithm estimates the sensor's response to illumination of the scene and corrects RAW data.

\textit{(e). Color Space Transformation} mainly includes two steps, first is mapping white balanced pixel to un-render color space (\textit{i.e.} CIEXYZ), and the second is mapping un-render color space to the display-referred color space (\textit{i.e.} sRGB), typically each use a 3$\times$3 matrix based on specific camera~\cite{Mobile_Computational}.


\textit{(f). Color and Tone Correction} are often implemented using 3D and 1D lookup tables (LUTs), while tone mapping also compresses pixel values.

\textit{(g). Sharpening}  enhances image details by unsharp masking or deconvolution.

We reference other steps like digital zoom and gamma correction from previous works~\cite{Michael_eccv16,Mobile_Computational,ISP_2005}. 
Meanwhile, in the ISP pipeline, many other operations prioritize the quality of the generated image rather than its performance in machine vision tasks.
Therefore, for specific adapter designs in \textbf{RAW-Adapter}, we selectively omit certain steps and focus on including the steps mentioned above. We provide detailed explanations in Sec.~\uppercase\expandafter{\romannumeral3}-A and Sec.~\uppercase\expandafter{\romannumeral3}-B (main paper).

\begin{table}[t]
\caption{Ablation analyze on RAW-Adapter's model structure.}
\label{tab:ablation_block}
\centering
\renewcommand\arraystretch{1.8}
\huge
\resizebox{8.8cm}{!}{
\begin{tabular}{c|ccccc|c|c|c}

\toprule
\toprule
\multirow{6}{*}{blocks} & \cellcolor{gray!20}base & \cellcolor{gray!20}$\mathbb{P_K}$ & \cellcolor{gray!20}$\mathbb{P_M}$ &  \cellcolor{gray!20}$\mathbb{L}$  &  \cellcolor{gray!20}$\mathbb{M}$  & \cellcolor{gray!20}mAP (Normal) & \cellcolor{gray!20}mAP (Overexposure $\bigstar$) & \cellcolor{gray!20}mAP (Low-light $\bigstar$)  \\ \cline{2-9} 
                        & \checkmark     &        &        &     &     &   89.2  &   88.8    & 82.6  \\
                        & \checkmark     & \checkmark   &        &     &     &  89.2 (+0.0)    &   88.8 (+0.0)  &  85.0 (+2.4) \\
                        & \checkmark     & \checkmark   & \checkmark   &     &     & 89.5 (+0.2) &   89.0 (+0.2)    &  86.2 (+3.6)     \\
                        & \checkmark     & \checkmark   & \checkmark   & \checkmark&     &   89.4 (+0.1)  &   89.0 (+0.2)    &  86.3 (+3.7) \\
                        & \checkmark     & \checkmark   & \checkmark   & \checkmark& \checkmark&   89.7 (+0.5)  & 89.5 (+0.7)    &  86.6 (+4.0) \\ \midrule \midrule
\multirow{6}{*}{blocks} & \cellcolor{gray!20}base & \cellcolor{gray!20}$\mathbb{P_K}$ & \cellcolor{gray!20}$\mathbb{P_M}$ &  \cellcolor{gray!20}$\mathbb{L}$  &  \cellcolor{gray!20}$\mathbb{M}$  & \cellcolor{gray!20}mAP (Fog $\clubsuit$) & \cellcolor{gray!20}mAP (Noise $\blacksquare$) & \cellcolor{gray!20}mAP (Vignetting $\blacksquare$)  \\ \cline{2-9} 
                        & \checkmark     &        &        &     &     &   79.4  &    88.0     &  86.4  \\
                        & \checkmark     & \checkmark   &        &     &     &  81.0 (+1.6)    &   87.9 (- 0.1)  &  87.6 (+1.2) \\
                        & \checkmark     & \checkmark   & \checkmark   &     &     & 81.3 (+1.9) &   88.5 (+0.5)   &  88.6 (+2.2)    \\
                        & \checkmark     & \checkmark   & \checkmark   & \checkmark&     &   81.2 (+1.8)  &   88.5 (+0.5)    &  89.0 (+2.6) \\
                        & \checkmark     & \checkmark   & \checkmark   & \checkmark& \checkmark&   81.4 (+2.0)  & 88.9 (+0.9)    &   89.3 (+2.9) \\ \bottomrule \bottomrule
\end{tabular}}
\vspace{-3mm}
\end{table}

\section{Ablation Analyze}
\label{sec2_Ablation}

\subsection{Impact of Different Blocks}

We conducted ablation experiments to evaluate the effectiveness of various components in RAW-Adapter. The experiments were carried out on the PASCAL(-D) dataset using RetinaNet~\cite{lin2017focal_loss} with a ResNet-50 backbone, and they covered both normal and various corrupted conditions. The results, presented in Table~\ref{tab:ablation_block}, show that the kernel predictor $\mathbb{P_K}$ delivers significant improvements in “Low-light $\bigstar$” (+2.4) and “Fog $\clubsuit$” (+1.6) scenes. These gains are mainly attributable to the enhancement provided by the gain ratio $g$ and denoising processes. However, $\mathbb{P_K}$ appears less effective in “Normal”, “Overexposure $\bigstar$”, and “Noise $\blacksquare$” scenes, possibly because current kernel-based denoising methods are too simplistic and may remove some fine details. Meanwhile, the implicit LUT $\mathbb{L}$ does not yield improvements under “Overexposure $\bigstar$” and “Low-light $\bigstar$” conditions but proves effective in the “Vignetting $\blacksquare$” condition. Finally, the model-level adapters $\mathbb{M}$ and the matrix predictor $\mathbb{P_M}$ contribute to performance improvements across all scenarios. Overall, each component in RAW-Adapter plays a unique role in addressing different conditions, and together they ensure robust performance.

\subsection{Different Parts in RAW-based Data Augmentation}

\begin{table}[t]
\caption{Ablation analyze on RAW-based data augmentation.}
\label{tab:ablation_aug}
\centering
\renewcommand\arraystretch{1.7}
\huge
\resizebox{8.8cm}{!}{
\begin{tabular}{c|cccc|c|c|c}

\toprule
\toprule

\multirow{6}{*}{blocks} & \cellcolor{gray!20}base & \cellcolor{gray!20}\textbf{B} & \cellcolor{gray!20}\textbf{C} & \cellcolor{gray!20}\textbf{Q}   & \cellcolor{gray!20}mAP (Low-light  $\bigstar$) & \cellcolor{gray!20}mAP (Low $\&$ Flare $\bigstar$) & \cellcolor{gray!20}mAP (Fog $\clubsuit$)  \\ \cline{2-8} 
                        & \checkmark     &    &    &     &  75.9  &  13.3   & 47.4 \\
                        &      & \checkmark   &        &     & 83.5 (+7.6)  & 20.8 (+7.5)  & 59.7 (+12.3)  \\
                        &     &   & \checkmark   &     & 78.7 (+2.8) & 16.6 (+3.3)  & 47.6 (+0.2) \\
                        &      &  &  & \checkmark&   76.2 (+0.3)  & 15.2 (+1.9) & 55.5 (+8.1)  \\
                        & \checkmark     & \checkmark   & \checkmark   & \checkmark&  84.9 (+9.0)  & 23.5 (+10.2)  & 61.8 (+14.4)  \\ \midrule \midrule
\multirow{6}{*}{blocks} & \cellcolor{gray!20}base & \cellcolor{gray!20}\textbf{B} & \cellcolor{gray!20}\textbf{C} &  \cellcolor{gray!20}\textbf{Q}   & \cellcolor{gray!20}mAP (Snow $\clubsuit$) & \cellcolor{gray!20}mAP (Defocus Blur $\spadesuit$) & \cellcolor{gray!20}mAP (CMOS-SD $\blacksquare$)  \\ \cline{2-8} 
                        & \checkmark     &  &    &   & 37.2 & 32.3  &  83.1 \\
                        &     & \checkmark   &        &     &  48.3 (+11.1) & 33.7 (+1.4) & 82.4 (-0.8) \\
                        &     &  & \checkmark   &     & 43.3 (+6.1) & 35.2 (+2.9) & 84.2 (+1.1)  \\
                        &   &  &   & \checkmark& 40.1 (+2.9) &  63.0 (+30.7) & 85.6 (+2.5) \\
                        & \checkmark     & \checkmark   & \checkmark   & \checkmark& 52.1 (+14.9) & 64.8 (+32.5) & 85.7 (+2.6) \\ \bottomrule \bottomrule
\end{tabular}}
\vspace{-3mm}
\end{table}

In addition to analyzing the different blocks in the RAW-Adapter design, we further examine how various components of RAW-based data augmentation (Section~\uppercase\expandafter{\romannumeral5} in the main paper) contribute to the RAW-Adapter's OOD performance. We present the results of RAW-Adapter on the PASCAL RAW-D dataset. As shown in Table.~\ref{tab:ablation_aug}, where $\textbf{B}$ denotes \textit{Brightness Variation}, $\textbf{C}$ denotes \textit{Chromaticity Augmentation} and $\textbf{Q}$ denotes \textit{Quality Degradation}. We observe that $\textbf{B}$ plays a crucial role in handling lightness degradations (“Low-light $\bigstar$” and “Low $\&$ Flare $\bigstar$”) and weather effects (“Fog $\clubsuit$” and “Snow $\clubsuit$”), but also causes the performance drop in “CMOS-SD  $\blacksquare$”. Meanwhile, $\textbf{C}$ improves performance across most scenarios, though the magnitude of improvement remains relatively limited compared with $\textbf{B}$ and $\textbf{Q}$. Additionally, Q is highly effective in both blurry scenes (“Defocus Blur  $\spadesuit$) and “CMOS-SD  $\blacksquare$”.

\section{Detailed Design of Input-level Adapters}
\label{sec3_Detail_Design}

In our main paper, we outlined that the input level adapters of RAW-Adapter comprise three components: the kernel predictor $\mathbb{P_K}$, the matrix predictor $\mathbb{P_M}$, and the neural implicit 3D LUT $\mathbb{L}$. In this section, we will provide a detailed explanation of how to set the parameter ranges for input-level adapters, along with conducting some results analysis. 

The kernel predictor $\mathbb{P_K}$ is responsible for predicting five ISP-related parameters, including the \ding{172} gain ratio $g$, the  Gaussian kernel \ding{173} $k$'s major axis radius $r_1$, \ding{174} $k$'s minor axis radius $r_2$, and the \ding{175} sharpness filter parameter $\sigma$.

\ding{172} The gain ratio $g$ is used to adjust the overall intensity of the image $\mathbf{I}_1$. Here $g$ initialized to 1 under normal light and over-exposure conditions. In low-light scenarios, $g$ is initialized to 5.

\ding{173} The major axis radius $r_1$ is initialized as 3, and we predict the bias of the variation of $r_1$, then add it to $r_1$.

\ding{174} The minor axis radius $r_2$ is initialized as 2, and we predict the bias of the variation of $r_2$, then add it to $r_2$.

\ding{175} The sharpness filter parameter $\sigma$ is constrained by a Sigmoid activation function to ensure its range is within (0, 1).

The matrix predictor $\mathbb{P_M}$ is responsible for predicting \ding{176} a white balance related parameter $\rho$ and \ding{177} white balance matrix  $\mathbf{E}_{ccm}$ (9 parameters). In total, 10 parameters need to be predicted.

\ding{176} $\rho$ is a hyperparameter of the Minkowski distance in SOG~\cite{Shades_of_gray} white balance algorithm. We set its minimum value to 1 and then use a ReLU activation function followed by adding 1 to restrict its range to (1, +$\infty$).

\ding{177} The matrix $\mathbf{E}_{ccm}$ 
  consists of the 9 parameters predicted by $\mathbb{P_M}$
  and forms a 3x3 matrix. No activation function needs to be added, it would directly added to the identity matrix $\mathbf{E}_{3}$ to form the final $\mathbf{E}_{ccm}$.

In the main text, we set the MLP dimension of the neural implicit 3D LUT (NILUT)~\cite{conde2024nilut} to 32 to balance performance and computational cost. Moving forward, in further we plan to explore more advanced architectures to further enhance LUT's effectiveness in RAW-based vision tasks.

\section{Segmentation on ADE20K-RAW (Synthesized)}
\label{sec4_ade20k_raw}

\begin{figure}
    \centering
    \includegraphics[width=0.92\linewidth]{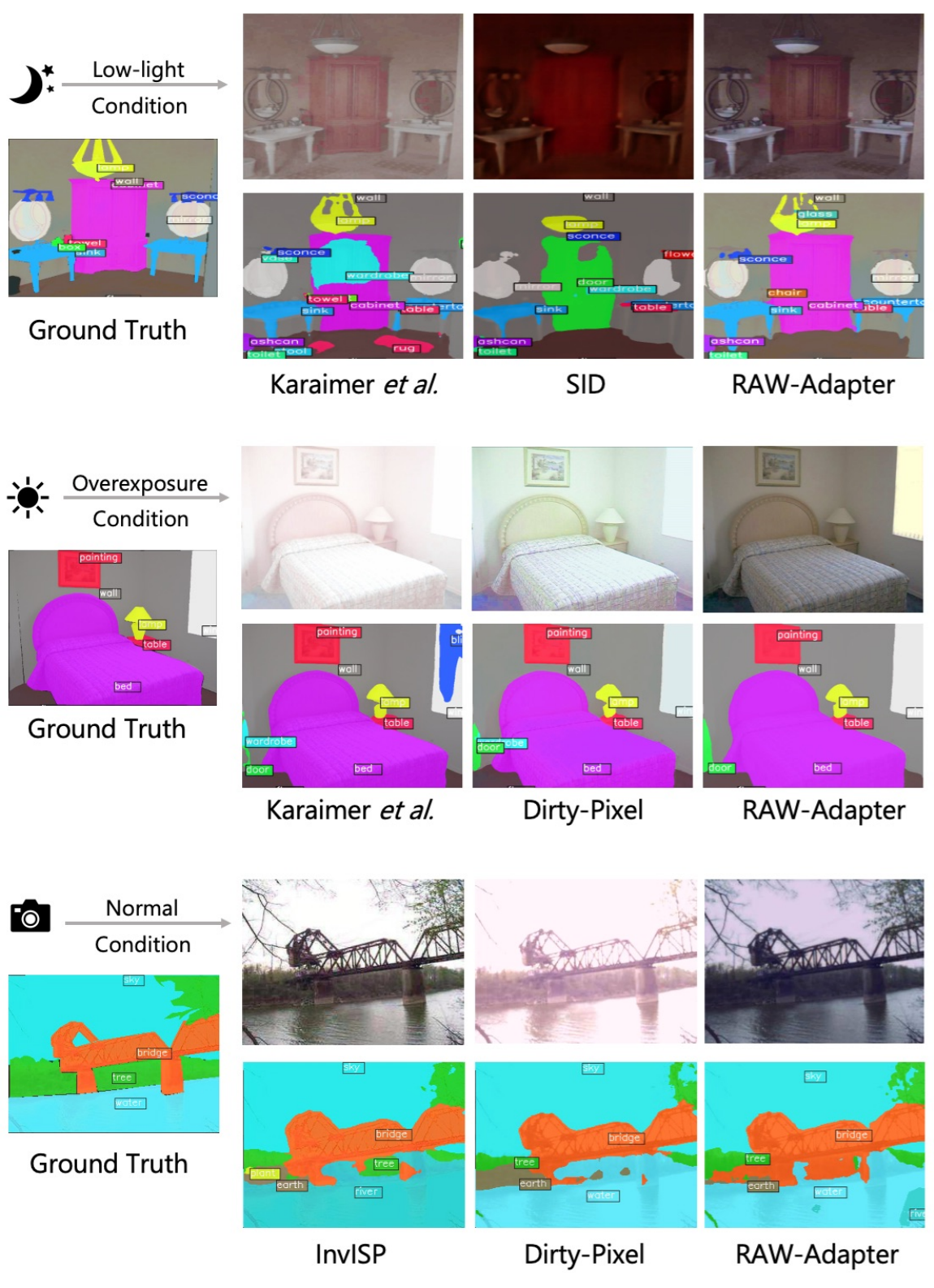}
    \vspace{-3mm}
    \caption{Visualization results on ADE20K-RAW~\cite{raw_adapter} dataset.}
    \label{fig:ade20kraw}
     \vspace{-3mm}
\end{figure}

\begin{table*}[t]
\caption{Comparison with other methods on \textbf{ADE 20K RAW} dataset (normal/overexposure/low-light).}
\vspace{-2mm}
\label{tab:segmentation}
\centering
\tiny
\renewcommand\arraystretch{1.2}
\resizebox{0.88\linewidth}{!}{%
\begin{tabular}{c|c|c|c|c|c|c}
\hline
\hline
  & backbone   &  params(M) $\downarrow$ & \begin{tabular}[c]{@{}c@{}}inference\\  time(s) $\downarrow$ \end{tabular} & \begin{tabular}[c]{@{}c@{}}mIOU $\uparrow$ \\ (Normal)\end{tabular} & \begin{tabular}[c]{@{}c@{}}mIOU $\uparrow$\\ (Overexposure)\end{tabular} & \  \begin{tabular}[c] {@{}c@{}}mIOU $\uparrow$\\ (Low-light)\end{tabular} \  \\ \hline
Demosacing  & \multirow{7}{*}{MIT-B5} &      \multirow{6}{*}{82.01}                                                     &    0.105   &  47.47   &  45.69    &    37.55        \\ \cline{1-1} \cline{4-7} 
Karaimer \textit{et al.}~\cite{Michael_eccv16} &                         &         &     0.525          & 45.48       & 42.85                                                    & 37.32                                               \\ \cline{1-1} \cline{4-7} 
InvISP~\cite{invertible_ISP}                                                                  &         &                & 0.203                                            & 47.82                                                   & 44.30                                                     & 4.03                                                \\ \cline{1-1} \cline{4-7} 
LiteISP~\cite{RAW-to-sRGB_ICCV2021}                                                                 &                         &                                                             &     0.261       & 43.22                                                   & 42.01                                                     & 5.52                                                 \\ \cline{1-1} \cline{4-7} 
DNF~\cite{jincvpr23dnf}                                                                     &                         &         &   0.186                                          & -                                                       & -                                                         & 35.88                                              \\ \cline{1-1} \cline{4-7} 
SID~\cite{SID}                                                                     &                         &      &    0.312                                       & -                                                       & -                                                         & 37.06 \\ \cline{1-1} \cline{3-7} 
\multirow{3}{*}{Dirty-Pixel~\cite{steven:dirtypixels2021}}     &   &  86.29  &   0.159  & \underline{47.86}  & \underline{46.50}  & 38.02 \\ \cline{2-7}  & \cellcolor[HTML]{ECF4FF}MIT-B3  & \cellcolor[HTML]{ECF4FF}48.92  & \cellcolor[HTML]{ECF4FF}0.098  & \cellcolor[HTML]{ECF4FF}46.19 & \cellcolor[HTML]{ECF4FF}44.13  & \cellcolor[HTML]{ECF4FF}36.93 \\ \cline{2-7} 
& \cellcolor{gray!10}MIT-B0 & \cellcolor{gray!10}8.00 & \cellcolor{gray!10}\underline{0.049}  & \cellcolor{gray!10}34.43 &  \cellcolor{gray!10}31.10  &      \cellcolor{gray!10}24.89                                                \\ \hline
\multirow{3}{*}{\begin{tabular}[c]{@{}c@{}}\textbf{RAW-Adapter} \\ (w/o $\mathbb{M}$)\end{tabular}}  & MIT-B5    &  82.09   &   0.148  &  47.83  &  46.48 & \underline{38.41} \\ \cline{2-7} & \cellcolor[HTML]{ECF4FF}MIT-B3   &   \cellcolor[HTML]{ECF4FF}44.72  & \cellcolor[HTML]{ECF4FF}0.086  & \cellcolor[HTML]{ECF4FF}46.22  & \cellcolor[HTML]{ECF4FF}44.00  &  \cellcolor[HTML]{ECF4FF}37.60  \\ \cline{2-7}    & \cellcolor{gray!10}MIT-B0  &  \cellcolor{gray!10}\textbf{3.80}  &  \cellcolor{gray!10}\textbf{0.032} &  \cellcolor{gray!10}34.66 &   \cellcolor{gray!10}31.82   & \cellcolor{gray!10}23.99 \\ \hline

\multirow{3}{*}{\textbf{RAW-Adapter}}  & MIT-B5    &  82.31   &  0.167  &  \textbf{47.95}  &  \textbf{46.62}  &  \textbf{38.75} \\ \cline{2-7} & \cellcolor[HTML]{ECF4FF}MIT-B3   &  \cellcolor[HTML]{ECF4FF}45.16   & \cellcolor[HTML]{ECF4FF}0.102  & \cellcolor[HTML]{ECF4FF}46.57  & \cellcolor[HTML]{ECF4FF}44.19  & \cellcolor[HTML]{ECF4FF}37.62   \\ \cline{2-7}    & \cellcolor{gray!10}MIT-B0       &  \cellcolor{gray!10}\underline{3.87}   & \cellcolor{gray!10}0.053  & \cellcolor{gray!10}34.72 & \cellcolor{gray!10}31.91  & \cellcolor{gray!10}25.06 \\ \hline

\multirow{3}{*}{\begin{tabular}[c]{@{}c@{}}\textbf{RAW-Adapter} \\ (w Augmentation)\end{tabular}}  & MIT-B5    &  82.31   &  0.167  &  \textbf{48.07}(+0.12)  &  \textbf{46.75}(+0.13)  &  \textbf{39.42}(+0.67) \\ \cline{2-7} & \cellcolor[HTML]{ECF4FF}MIT-B3   &  \cellcolor[HTML]{ECF4FF}45.16   & \cellcolor[HTML]{ECF4FF}0.102  & \cellcolor[HTML]{ECF4FF}46.91(+0.34)  & \cellcolor[HTML]{ECF4FF}44.54(+0.35)  & \cellcolor[HTML]{ECF4FF}38.04(+0.42)   \\ \cline{2-7}    & \cellcolor{gray!10}MIT-B0       &  \cellcolor{gray!10}\underline{3.87}   & \cellcolor{gray!10}0.053  & \cellcolor{gray!10}34.85(+0.13) & \cellcolor{gray!10}32.07(+0.16)  & \cellcolor{gray!10}27.33(+2.27) \\
\hline
\hline
\end{tabular}}
\vspace{-2mm}
\end{table*}

In our conference version~\cite{raw_adapter}, we synthesized the ADE20K-RAW dataset using InvISP on the ADE20K dataset and set up three conditions: normal, low-light, and overexposure. We compared our method with previous ISP algorithms~\cite{SID,jincvpr23dnf,invertible_ISP,RAW-to-sRGB_ICCV2021,Michael_eccv16} as well as joint-training approach~\cite{steven:dirtypixels2021}. Using the SegFormer~\cite{xie2021segformer} framework with three backbone weights (MIT-B0, MIT-B3, and MIT-B5), the results are presented in Table~\ref{tab:segmentation}. Additionally, we incorporated RAW-based data augmentation and found that it significantly improved performance on ADE20K-RAW, further validating the effectiveness of our augmentation strategy. The visualization results are shown in Fig.~\ref{fig:ade20kraw}.

\section{More Visualization Results}
\label{sec5_visulization}

\begin{figure*}
    \centering
    \includegraphics[width=1.0\linewidth]{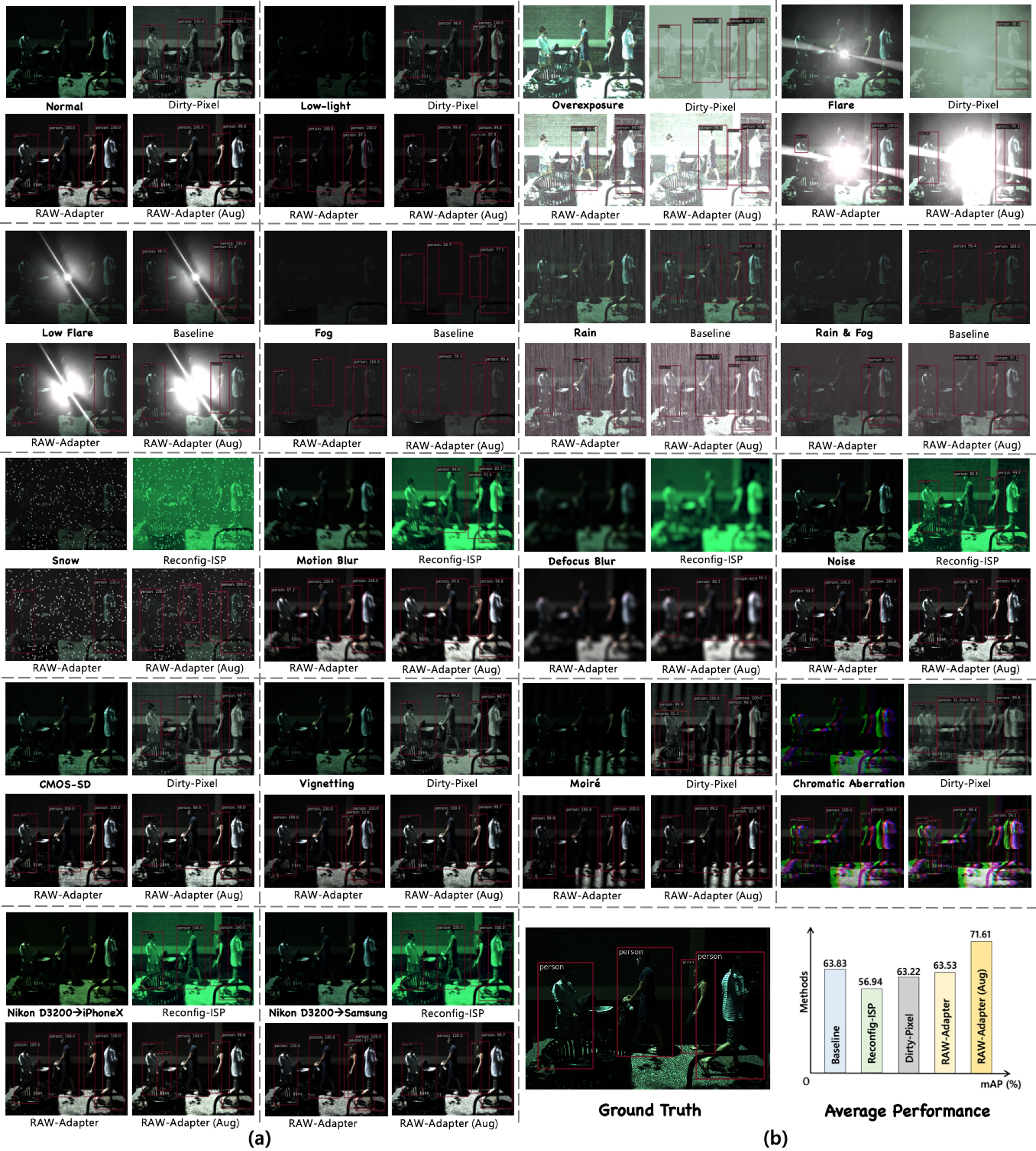}
    \caption{(a) Out-of-domain (OOD) object detection results on the PASCAL RAW-D dataset, comparing with original baseline (direct RAW input), Dirty-Pixel~\cite{steven:dirtypixels2021} and Reconfig-ISP~\cite{yu2021reconfigisp}. (b) Ground truth and average OOD mAP ($\%$) results.}
    \label{fig:supp_det}
\end{figure*}
\begin{figure*}
    \centering
    \includegraphics[width=1.0\linewidth]{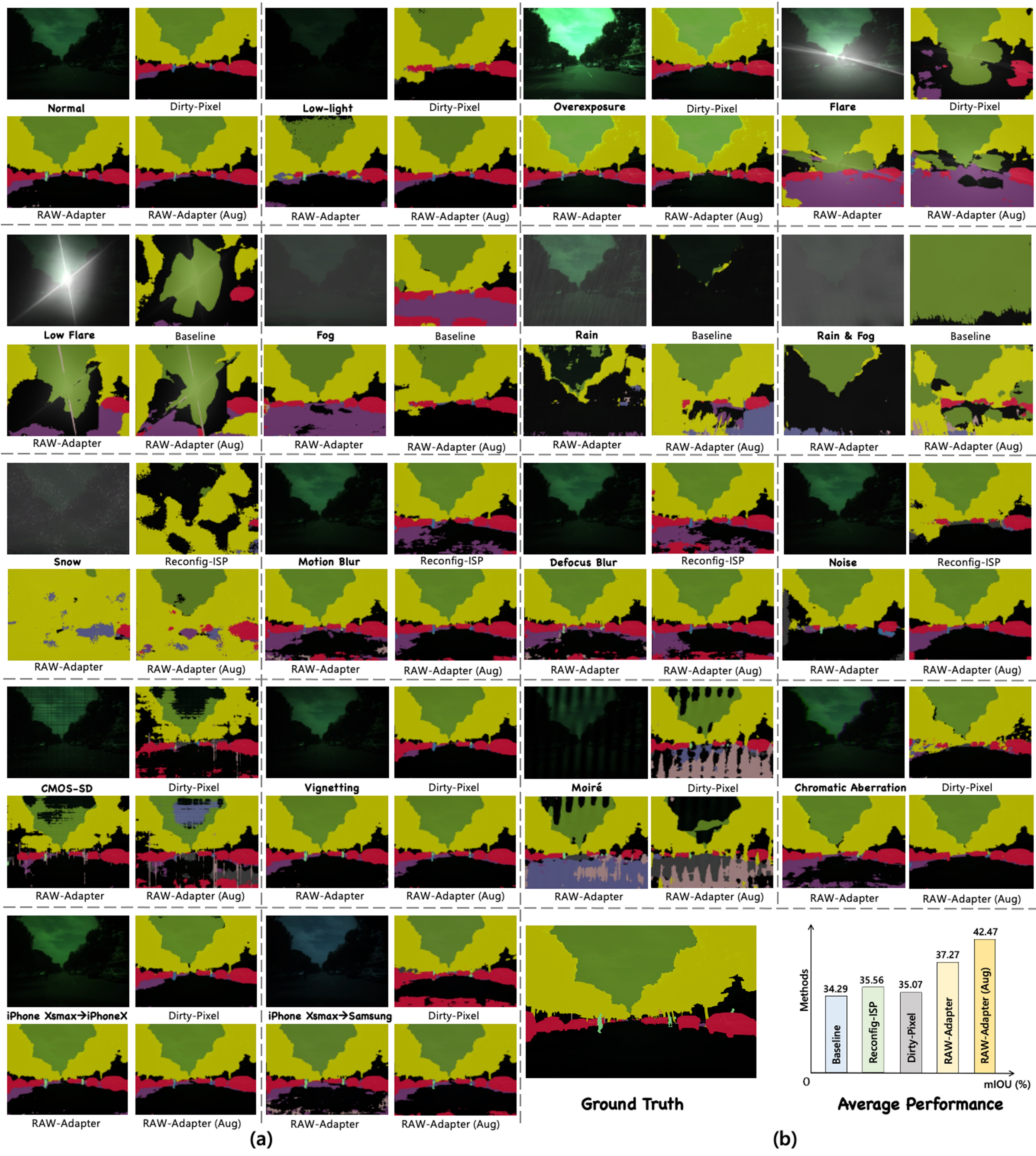}
    \caption{(a) Out-of-domain (OOD) semantic segmentation results on the iPhone XS Max-D dataset, comparing with original baseline (direct RAW input), Dirty-Pixel~\cite{steven:dirtypixels2021} and Reconfig-ISP~\cite{yu2021reconfigisp}. (b) Ground truth and average mIOU ($\%$) results.}
    \label{fig:supp_seg}
\end{figure*}

We present additional object detection visualization results in Fig.~\ref{fig:supp_det} (corresponding to Table.~\uppercase\expandafter{\romannumeral6} in the main paper) and more semantic segmentation visualization results in Fig.~\ref{fig:supp_seg} (corresponding to Table.~\uppercase\expandafter{\romannumeral7} in the main paper). The background images produced by the different methods represent the outcomes after encoder enhancement. Notably, the backgrounds in images enhanced by RAW-Adapter exhibit the most harmonious colors, whereas Reconfig-ISP tends to shift backgrounds towards a greenish hue when handling OOD cases. Furthermore, RAW-based data augmentation significantly improves OOD perception performance. For example, the ``Rain $\&$ Fog $\clubsuit$'' and ``Low $\&$ Flare $\bigstar$'' conditions in Fig.~\ref{fig:supp_det} for object detection, as well as the ``Fog $\clubsuit$'' and ``Noise $\blacksquare$'' conditions in Fig.~\ref{fig:supp_seg} for semantic segmentation, clearly illustrate these improvements.

\vfill

\end{document}